\documentclass{article}

\usepackage{microtype}
\usepackage{graphicx}
\usepackage{subcaption}
\usepackage{dsfont}
\usepackage{booktabs} %

\usepackage{amsmath}
\usepackage{amsfonts}
\usepackage{bm}
\usepackage{amsthm}
\usepackage{color}
\usepackage{wrapfig}
\usepackage{cancel}
\usepackage{algorithm}
\usepackage{algorithmic}
\usepackage{multirow}
\renewcommand{\algorithmicrequire}{\textbf{Input:}}
\renewcommand{\algorithmicensure}{\textbf{Output:}}

\usepackage{hyperref}
\usepackage{cleveref}
\usepackage{makecell}

\newcommand{\inbatch}{\mathcal{M}}
\newcommand{\bx}{\mathbf{x}}
\newcommand{\bbm}{\mathbf{m}}
\newcommand{\bV}{\mathbf{V}}

\newcommand{\bk}{\mathbf{k}}

\newcommand{\bX}{\mathbf{X}}

\newcommand{\bZ}{\mathbf{Z}}
\newcommand{\bU}{\mathbf{U}}
\newcommand{\by}{\mathbf{y}}
\newcommand{\bz}{\mathbf{z}}

\newcommand{\bw}{\mathbf{w}}

\newcommand{\bs}{\mathbf{s}}

\newcommand{\bbf}{\mathbf{f}}

\newcommand{\bu}{\mathbf{u}}
\newcommand{\bW}{\mathbf{W}}

\newcommand{\bmu}{\bm{\mu}}
\newcommand{\bSigma}{\bm{\Sigma}}
\newcommand{\bI}{\mathbf{I}}
\newcommand{\bzero}{\mathbf{0}}
\newcommand{\bK}{\mathbf{K}}

\newcommand{\bphi}{\bm{\phi}}

\newcommand{\bOmega}{\bm{\Omega}}

\newcommand{\natgrad}{\tilde{\nabla}}

\DeclareMathOperator*{\argmax}{arg\,max}

\newcommand{\bigO}{\mathcal{O}}

\newcommand{\KL}[2]{\mathrm{KL}\left[#1\|#2\right]}

\newtheorem{rem}{Remark}

\usepackage[accepted]{icml2019}

\icmltitlerunning{Scalable Training of Inference Networks for Gaussian-Process Models}

\begin{document}
    
    \twocolumn[
    \icmltitle{Scalable Training of Inference Networks for Gaussian-Process Models}

    \icmlsetsymbol{equal}{*}
    
    \begin{icmlauthorlist}
        \icmlauthor{Jiaxin Shi}{tsinghua}
        \icmlauthor{Mohammad Emtiyaz Khan}{riken-aip}
        \icmlauthor{Jun Zhu}{tsinghua}
    \end{icmlauthorlist}
    
    \icmlaffiliation{tsinghua}{Dept. of Comp. Sci. \& Tech., Institute for AI, BNRist Center, THBI Lab, Tsinghua University, Beijing, China}
    \icmlaffiliation{riken-aip}{RIKEN Center for Advanced Intelligence project, Tokyo, Japan}
    
    \icmlcorrespondingauthor{Jiaxin Shi}{shijx15@mails.tsinghua.edu.cn}
    \icmlcorrespondingauthor{Jun Zhu}{dcszj@tsinghua.edu.cn}
    
    \icmlkeywords{Machine Learning, ICML}
    
    \vskip 0.3in
    ]

    \printAffiliationsAndNotice{}  %
    
    \begin{abstract}
        Inference in Gaussian process (GP) models is computationally challenging for large data, and often difficult to approximate with a small number of inducing points.
        We explore an alternative approximation that employs stochastic inference networks %
        for a flexible inference.
        Unfortunately, for such networks, minibatch training is difficult to be able to learn meaningful correlations over function outputs for a large dataset.
        We propose an algorithm that enables such training by tracking a stochastic, functional mirror-descent algorithm.
        At each iteration, this only requires considering a finite number of input locations, resulting in a scalable and easy-to-implement algorithm. 
        Empirical results show comparable and, sometimes, superior performance to existing sparse variational GP methods.
    \end{abstract}
    \vspace{-0.4cm}
    \section{Introduction}
    \label{sec:intro}
    
    Gaussian processes (GP)~\citep{rasmussen2006gaussian} and their deep variants \citep{damianou2013deep} are powerful nonparametric distributions for both supervised \citep{williams1996gaussian, bernardo1998regression, williams1998bayesian} and unsupervised machine-learning \citep{lawrence2005probabilistic, damianou2016variational}. Such processes can generate smooth functions to model complex data and provide principled approaches for uncertainty quantification. Despite this, their application has been limited because Bayesian inference for such modeling requires inversion of a matrix which is computationally challenging (typically $\bigO(N^3)$ for $N$ data examples).
    
    Many methods have been proposed to tackle this issue, and they mostly rely on finding a small number of \emph{inducing points} in the input\footnote{For supervised learning, these can be features, while for unsupervised learning, these could be latent vectors.} space to reduce the cost of matrix inversion \cite{quinonero2005unifying, titsias2009variational}.
    Methods that employ variational inference to estimate inducing points \citep{titsias2009variational} can scale well to large data by minibatch stochastic-gradient methods~\citep{hensman2013gaussian}, but the quality of posterior approximations obtained could be limited since the number of inducing points needs to be small, even when the data are extremely large~\citep{cheng2017variational}. 
    Optimization of inducing points is another challenging problem when the objective is nonconvex and stochastic-gradient methods can get stuck~\citep{bauer2016understanding}.
    
    \begin{figure}[t]
        \centering
        \includegraphics[width=\linewidth,trim=1cm 4cm 1cm 2cm,clip]{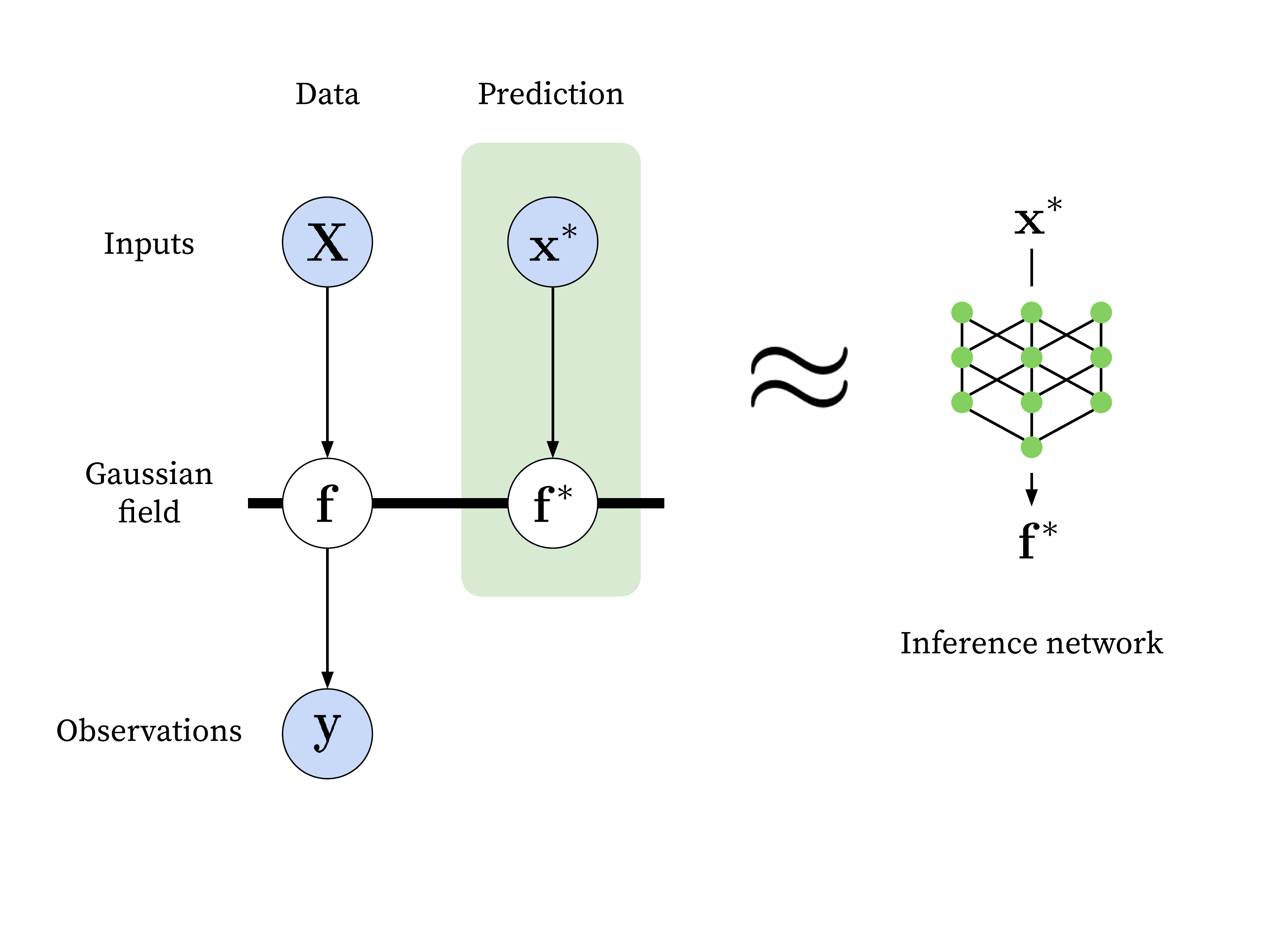}
        \caption{Inference networks for GPs are an alternative way to predict output $\bbf^*$ given test inputs $\bx^*$.}
        \label{fig:infer-net}
        \vspace{-0.6cm}
    \end{figure}
    
    A relatively less-explored approach is to approximate the GP posterior distribution by using function approximators, such as deep neural networks~\citep{sun2018functional}. By introducing randomness in the parameters of such \emph{inference networks}, it is possible to generate functions similar to those generated by the posterior distribution, as illustrated in Fig.~\ref{fig:infer-net}.
    By directly approximating the functions, we can obtain more flexible alternatives to sparse variational approaches.   
    Unfortunately, unlike sparse methods, training with minibatches is challenging in the function space. 
    The difficulty is to be able to maintain meaningful correlations over function outputs while only looking at a handful of examples in each iteration.
    Scalable training of such inference networks is extremely important for them to be useful for flexible posterior inference in GP models. 
    
    In this paper, we propose an algorithm to scalably train the network by tracking an adaptive Bayesian filter defined in the function space. The filter is obtained by using a stochastic, functional mirror-descent algorithm which is guaranteed to converge to the exact posterior process but is computationally intractable. 
    By bootstrapping from approximations given by inference networks, we can use stochastic gradients to train the network with minibatches of data.
    We demonstrate training of various types of networks, such as those based on random feature expansions and deep neural networks.
    Our results show that the problem caused by incorrectly minibatching in previous works is fixed by our method. 
    Unlike exist works that use neural networks to parameterize deep kernel in the GP prior~\citep{wilson2016deep}, our variational treatment prevents overfitting and allows arbitrarily complex networks, which often achieve better performance than inducing-point approaches in regression tasks.
    Finally, we show that our method is a more flexible alternative for GP inference than existing sparse methods, which allows us to train deep ConvNets for the inference of GPs induced from infinite-width Bayesian ConvNets.
    
    \vspace{-0.1cm}
    \subsection{Related work}
    \vspace{-0.05cm}
    For a scalable inference in GP models, plenty of work has been done on developing sparse methods \citep{quinonero2005unifying, titsias2009variational, hensman2013gaussian}.
    These works have made GPs a viable choice for practical problems by making minibatch training possible.
    Our work takes a different approach than these methods by using inference networks for posterior approximation.
    The computation complexity of our method is similar to sparse methods, but our posterior approximations are much more flexible.
    Currently, it is challenging to train such inference networks in the function space while achieving a good performance.
    Our work fills this gap and shows that it is indeed possible to scalably train them and get a similar, and sometimes better, performance than sparse GP methods.
    
    A few recent approaches have used inference networks for posterior approximations in the function space, although they have not directly applied it to inference in GP models.
    For example, Neural Processes (NP)~\citep{garnelo2018conditional,garnelo2018neural} uses inference networks on a model with a learned prior, while Variational Implicit Processes (VIP)~\citep{ma2018variational}, in a similar setup, uses a GP for approximations.
    \citet{hafner2018reliable} design a function-space prior called the noise contrastive prior.
    The work of \citet{sun2018functional} is perhaps the closest to our approach, but they use an heuristic procedure for minibatch training.
    Our work proposes a scalable minibatch training method which is useful for all of these existing works.
    
    The idea of tracking a learning process to \emph{distill} information has been used in many previous approaches, e.g., \citet{bui2017streaming} use streaming variational Bayes for inference in sparse GPs, while \citet{balan2015bayesian} track stochastic gradient Langevin dynamics (SGLD) (the \emph{teacher}) to distill information into a neural network (the \emph{student}). Our method can also be interpreted in a teacher-student framework, where the teacher can be obtain from the student network by taking a stochastic mirror descent step, which is much cheaper than simultaneously running another inference algorithm like SGLD.
    
    Some of our inference networks are based on random feature expansion and are related to existing work on spectrum approximations for GPs \citep{lazaro2010sparse, gal2015improving, hensman2017variational}.  Among them \citet{lazaro2010sparse} does function-space inference and is similar to our work, but is a full-batch algorithm.  %
    
    \section{Bayesian Inference in GP Models}
    We start by discussing the challenges associated with existing methods for inference in GP models, and then discuss the difficulty in scalable training of inference networks.
    
    \vspace{-0.1cm}
    \subsection{Gaussian Processes}
    \vspace{-0.05cm}
    \label{sec:intro-gp}
    A Gaussian Process (GP) is a stochastic process defined using a mean function $m(\bx)$ and covariance function $k(\bx, \bx')$:
    \begin{equation}
    f(\bx)\sim \mathcal{GP}\left(m(\bx), k(\bx, \bx')\right).
    \end{equation}
    A remarkable property of GPs is that, for any finite number of inputs $\bX = \left[\bx_1, \dots, \bx_N\right]^\top$, the marginal distribution of the function values $\bbf = \left[f(\bx_1), \dots, f(\bx_N)\right]^\top$ follow a  multivariate Gaussian distribution: $
    \bbf \sim \mathcal{N}\left( \mathbf{m}_{\mathcal{D}}, \bK_{\mathcal{D},\mathcal{D}}\right),  $
    where $\mathcal{D} := [1,2,\dots,N]$ are the indices of the data examples, $\mathbf{m}_{\mathcal{D}}$ is a vector with entries $m(\bx_i)$ and  $\bK_{\mathcal{D},\mathcal{D}}$ is a matrix with $(i,j)$'th entry as $k(\bx_i, \bx_j)$. This property can be utilized to obtain function approximations for the modeling of complex data. 
    For example, for regression analysis given input features $\bX$, the function values $\bbf$ can be used to model the output mean: $\by \sim \mathcal{N}(\bbf, \sigma^2 \bI)$ where $\sigma^2$ is the noise variance. Correlations in the data vector $\by$ can now be explained through the correlation defined using the GP prior in the function space.
    This approach is widely used to design nonlinear methods for supervised and unsupervised learning.
    
    Given such a prior distribution, the goal of Bayesian inference is to compute the posterior distribution over the function $f(\bx)$ evaluated at arbitrary test inputs $\bx$. 
    For Gaussian likelihoods such as in regression, due to the conjugacy, the posterior distribution takes a convenient closed-form solution, thus the predictive distribution at a test location $\bx^*$ is available in closed-form:
    \begin{align}
    f(\bx^*)|\by \sim \mathcal{N}&\left(\bk_{*,\mathcal{D}}^T(\bK_{\mathcal{D},\mathcal{D}} + \sigma^2\bI)^{-1}(\by-\mathbf{m}_{\mathcal{D}}), \right.\notag \\
    &\quad\left.k_{**} - \bk_{*\mathcal{D}}^T(\bK_{\mathcal{D},\mathcal{D}} + \sigma^2\bI)^{-1}\bk_{\mathcal{D},*}\right),
    \label{eq:gpreg}
    \end{align}
    where $[\bk_{*\mathcal{D}}]_{i} = k(\bx^*, \bx_i)$ and $k_{**} = k(\bx^*, \bx^*)$. 
    \Cref{eq:gpreg} is difficult to compute in practice when $N$ is large. The computational cost of matrix inversion %
    is in $\bigO(N^3)$. 
    This inversion is only possible when $N$ is of moderate size, usually only a few thousands.
    
    \subsection{Sparse Approximations for GPs}
    \label{sec:intro-svgp}
    
    Sparse-approximation methods reduce the computation cost by choosing a small number of $M$ points, where $M\ll N$, to perform the matrix inversion. These points are known as the \emph{inducing points}.
    A variety of such methods have been proposed~\citep{quinonero2005unifying} and they mostly differ in the manner of selection of these points and the kind of approximations used for the matrix inverse.
    Among these, methods based on variational inference are perhaps one of the most popular~\citep{titsias2009variational}.
    Given $M$ inducing points $\bZ := \left[\bz_1, \bz_2, \dots, \bz_M\right]^\top$ and their function values $\bu := \left[f(\bz_1),f(\bz_2),\dots,f(\bz_M) \right]^\top$, these methods approximate the posterior distribution $p(\bbf,\bu|\by, \bX, \bZ)$ with a variational distribution $q(\bbf,\bu)$.
    By restricting the variational approximation to be $q(\bbf, \bu) := q(\bu)p(\bbf|\bu)$, the variational lower bound is greatly simplified, 
    \begin{equation} \label{eq:sparse-lb}
    \mathcal{L}(q, \bZ) := \mathbb{E}_{q(\bu)p(\bbf|\bu)} \left[ \log p(\by|\bbf) \right] - \mathrm{KL}[q(\bu)\|p(\bu)].
    \end{equation}
    For GP regression, the optimal $q(\bu)$ is a Gaussian whose parameters can be obtained in closed-form~\citep{titsias2009variational}.
    Even though the computation of $q(\bu)$ scales linearly with $N$, its covariance matrix can be inverted in $\bigO(M^3)$ which reduces the prediction cost drastically.
    The linear dependence on $N$ can be further reduced by using a minibatch training proposed in \citet{hensman2013gaussian}, which makes the method scale well to large data. Sparse variational methods can scale well, and also perform reasonably with enough number of properly chosen inducing-points.
    
    In practice, however, the number of inducing points is limited to make the matrix inversion feasible. This limits the flexibility of the posterior distribution whose complexity might grow with the complexity of the problem.
    Finding good inducing points is another challenging issue. The landscape of the lower bound with respect to $\bZ$ presents a difficult optimization problem, and, so far, there are no good methods for this problem.
    
    \subsection{Function Space View and Inference Networks}
    \label{sec:gp-func}
    
    The sparse variational method discussed above can be derived by using a function-space view of GPs.
    As discussed in several recent works \citep{cheng2016incremental, cheng2017variational, mallasto2017learning}, a GP has a dual representation in a separable Banach space, which contains the RKHS $\mathcal{H}$ induced by the covariance kernel. This view motivates to directly apply variational inference in the function space. 
    Following \citet{cheng2016incremental}, if we denote the dual representation of the posterior and variational distributions by $p(f|\by)$ and $q(f)$ respectively, then the variational objective in the function space can be written as following:
    \begin{align} 
    \label{eq:felbo}
    \mathcal{L}(q(f)) &= \mathbb{E}_{q(f)} \left[ \log p(\by|f) \right] - \KL{q(f)}{p(f)}.
    \end{align}
    We recover the sparse variational GP problem in \eqref{eq:sparse-lb} when we restrict $q(f) \propto p(f_*|\bu)q(\bu)|\bK_{\bu}|^{1/2}|\bK_* - \bK_{*\bu}\bK_{\bu}^{-1}\bK_{\bu*}|^{1/2}$, where $f_*$ represents the function outputs not covered by $\bu$. The two determinants arise from the change of measure from function $f$ to its outputs:%
    $f(\bx) = \langle f, k(\bx, \cdot)\rangle_{\mathcal{H}}$.
    
    We can improve the flexibility of the approximations by employing better choices of $q(f)$.
    Function approximators with random parameters, such as stochastic neural networks, are such alternatives, where we can learn to generate functions that mimic the samples from the posterior. This is illustrated in Fig.~\ref{fig:infer-net}.
    Drawing an analogy to the networks used for inference in deep generative models~\cite{kingma2013auto}, we call them \emph{inference networks}.
    When trained well, such networks can yield much more flexible posterior approximations than sparse variational approximation which is restricted by the choice of inducing points.
    
    Unfortunately, training inference networks is much more challenging than sparse methods.
    The sparse approach simplifies the problem to a parametric form where we only need to deal with a small number inducing points and data examples.
    It is challenging to design a similar procedure without any sparse assumption on $q(f)$.
    
    Existing approaches have mostly relied on heuristic procedures to solve this problem. 
    For example, a recent approach of \citet{sun2018functional} proposes to match marginal distributions of $q(f)$ and $p(f|\by)$ at a finite number of \emph{measurement points} $\bX_\inbatch := [\bx_1,\bx_2,\dots,\bx_M]^\top$ sampled from a distribution $c(\bx)$, by minimizing\footnote{We drop the dependence of $p$ and $q$ on training data $\bX$ and sampled locations $\bX_\inbatch$.} $\KL{q(\bbf_\inbatch)}{p(\bbf_\inbatch|\by)}$ where $\bbf_\inbatch$ are the function values at $\bX_\inbatch$.
    This is a reasonable criterion to match two GPs, since they are completely determined by their first two moments and the solution is unique when $M>2$~\citep{sun2018functional}. %
    However, this is challenging to optimize due to the dependence of the term $p(\bbf_\inbatch|\by)$ on the whole dataset $\by$. 
    \citet{sun2018functional} propose to pick subsets of data at each iteration but then the minimization problem no longer corresponds to matching $q(f)$ and $p(f|\by)$ faithfully.
    Many other existing works on function-space inference with neural networks face similar challenges when it comes to minibatch training~\citep{wang2018function,ma2018variational, hafner2018reliable}.
    Such scalable training of inference network, while maintaining meaningful correlations in function outputs for a large dataset, remains a challenging problem.
    
    \section{Scalable Training of Inference Networks}
    \label{sec:scalable_infnet}
    We present a new algorithm to scalably train inference networks by using minibatches. Our main idea is to track a stochastic, functional mirror descent algorithm to enable efficient minibatch training. We start with a brief description of the mirror descent algorithm. 
    
    \vspace{-0.1cm}
    \subsection{Stochastic Functional Mirror Descent Algorithm}
    \vspace{-0.05cm}
    \label{sec:smd}
    
    We follow the functional mirror-descent method proposed in \citet{dai2016provable,cheng2016incremental} to optimize \eqref{eq:felbo}. 
    The functional mirror descent algorithm is an extension of gradient descent where gradient steps are taken in a function space and the length of the steps is measured by using a Bregman divergence (e.g., KL) instead of a Euclidean distance.
    A stochastic version of this algorithm is analogous to stochastic gradient descent where a minibatch of data could be used to build a stochastic approximation of the functional gradient.
    The method %
    takes the following form:
    \begin{equation} \label{eq:fsmd}
    \resizebox{.88\columnwidth}{!}{$\displaystyle q_{t+1} = \argmax_{q} \int \hat{\partial}\mathcal{L}(q_t)q(f) df - \frac{1}{\beta_t}\KL{q}{q_{t}}.$}
    \end{equation}
    where $t$ is the iteration, $\beta_t>0$ is the learning rate, $q_t:=q_t(f)$ is the previous approximation, and $\hat{\partial}\mathcal{L}(q_t) = N\log p(y_n|f) + \log p(f) - \log q_t(f)$ is an unbiased stochastic approximation of the functional gradient of $\mathcal{L}(q)$ at $q=q_t$ obtained by randomly sampling a data example $n$. 
    An attractive property is that there is a closed-form solution given as follows,
    \begin{equation} \label{eq:frac-bayes}
    q_{t+1}(f) \propto p(y_n|f)^{N\beta_t} p(f)^{\beta_t}q_t(f)^{1 - \beta_t}.
    \end{equation}
    This update can be seen as an \emph{adaptive Bayesian filter} where the previous posterior approximation $q_t(f)$ is used to modify the prior distribution $p(f)$ and a likelihood of the subsampled data is used to update the posterior approximation. We note that the functional mirror descent is equivalent to natural gradient for exponential family densities. In \cref{app:natgrad} we provide an alternative derivation of \eqref{eq:frac-bayes} by taking natural gradient on the exponential family representation of GP.
    
    Each step of this algorithm only requires subsampling a single data point, which makes it attractive for our purposes. The algorithm is also guaranteed to converge to the true posterior $p(f|\by)$ as discussed in \citet{dai2016provable}, %
    where a particle-based approach is proposed for the update in \eqref{eq:frac-bayes}. \citet{khan2017conjugate} used a parametric version where the update can be performed analytically.
    Our case is different from these works because the random variable is a function, thus infinite-dimensional and difficult to be represented in a compact form.
    We instead use an inference network to implement this filter. %
    We will use a version of \eqref{eq:frac-bayes} to compute stochastic gradients and update the parameters of the inference network. This is explained next.
    
    \subsection{Minibatch Training of Inference Networks}
    
    We propose a tractable approximation to \eqref{eq:frac-bayes} by bootstrapping from the inference network at each iteration.
    We denote the inference network by $q_\gamma(f)$ where $\gamma$ is the set of parameters that we wish to estimate.
    We assume that we can evaluate the network at a finite set of $M$ points $\bX_\inbatch := [\bx_1, \bx_2,\dots,\bx_M]^\top$ and that, just like a GP, the evaluated points follow a Gaussian distribution, i.e., $
    q_{\gamma}(\bbf_{\inbatch}) = \mathcal{N}( \bmu_{\inbatch}, \bSigma_\inbatch), $
    where $\bmu_\inbatch$ and $\bSigma_\inbatch$ are the mean and covariance that depend on the parameter $\gamma$. In \cref{sec:infer-net}, we will give many examples of inference networks that have this property.
    
    With such a $q_\gamma(f)$, we hope to track \eqref{eq:frac-bayes}, so that $q_{\gamma_{t+1}}(f)$ moves closer to the true posterior process than $q_{\gamma_t}(f)$.
    For this purpose, 
    an obvious solution is to replace $q_t(f)$ by $q_{\gamma_t}(f)$, i.e., %
    we can bootstrap from the current posterior approximation given by the inference network:
    \begin{equation} \label{eq:frac-bayes-approx}
    \hat{q}_{t+1}(f) \propto p(y_n|f)^{N\beta_t} p(f)^{\beta_t} {\color{red} q_{\gamma_t}(f)}^{1 - \beta_t}.
    \end{equation}
    The idea of bootstrap has long been used in particle filtering~\citep{doucet2001introduction} to obtain better posterior approximations.
    An attractive property of \eqref{eq:frac-bayes-approx} for GP regression is that, given inputs $\bX_\inbatch$, all the quantities in the right hand side follow a Gaussian distribution, therefore $\hat{q}_{t+1}$ has a Gaussian distribution whose mean and covariance are available in closed-form. 
    The last two terms in \eqref{eq:frac-bayes-approx} are Gaussian and can be multiplied to get the new GP prior:
    \begin{align}
    & p(\bbf_\inbatch, f_n)^{\beta_t}  q_{\gamma_t}(\bbf_\inbatch, f_n)^{1 - \beta_t} \nonumber \\
    & := 
    \mathcal{N}\left(%
    \left[%
    \begin{array}{c}
    \widetilde{\bbm}_\inbatch\\ \widetilde{\bbm}_n 
    \end{array}
    \right] ,
    \left[%
    \begin{array}{cc}
    \widetilde{\bK}_{\inbatch,\inbatch} & \widetilde{\bK}_{\inbatch,n} \\
    \widetilde{\bK}_{n,\inbatch} & \widetilde{\bK}_{n,n} 
    \end{array}
    \right]
    \right), \\
    &\propto \quad  \mathcal{N}\left(
    \left[%
    \begin{array}{c}
    0\\ 0 
    \end{array}
    \right] ,
    \left[%
    \begin{array}{cc}
    \bK_{\inbatch,\inbatch} & \bK_{\inbatch,n} \\
    \bK_{n,\inbatch} & \bK_{n,n} 
    \end{array}
    \right]
    \right)^{\beta_t} \nonumber \\
    &\quad
    \times 
    \mathcal{N}\left(%
    \left[%
    \begin{array}{c}
    \bmu_\inbatch\\ \bmu_n 
    \end{array}
    \right] ,
    \left[%
    \begin{array}{cc}
    \bSigma_{\inbatch,\inbatch} & \bSigma_{\inbatch,n} \\
    \bSigma_{n,\inbatch} & \bSigma_{n,n} 
    \end{array}
    \right]
    \right)^{(1-\beta_t)} , 
    \end{align}
    where $\widetilde{\bbm}$ and $\widetilde{\bK}$ denotes the new mean and covariance. Using this in \eqref{eq:frac-bayes-approx} and multiplying by the Gaussian likelihood, we get the following new GP regression problem, which has a closed-form solution similar to \eqref{eq:gpreg}: 
    \begin{align}
    & \hat{q}_{t+1}(\bbf_\inbatch, f_n) \propto \mathcal{N}(y_n|f_n, \sigma^2/(N\beta_t) ) \nonumber \\
    &\quad
    \times 
    \mathcal{N}\left(%
    \left[%
    \begin{array}{c}
    \widetilde{\bbm}_\inbatch\\ \widetilde{\bbm}_n 
    \end{array}
    \right] ,
    \left[%
    \begin{array}{cc}
    \widetilde{\bK}_{\inbatch,\inbatch} & \widetilde{\bK}_{\inbatch,n} \\
    \widetilde{\bK}_{n,\inbatch} & \widetilde{\bK}_{n,n} 
    \end{array}
    \right]
    \right).
    \label{eq:frac-bayes-approx-gpreg}
    \end{align}
    The marginal $\hat{q}_{t+1}(\bbf_\inbatch)$ can be read from this directly. Though we have access to any finite marginal distribution of $\hat{q}_{t+1}(f)$, mapping this to the inference network parameters is difficult.
    We can use the approach of \citet{sun2018functional} to match the marginals of the $q_{\gamma}(f)$ and $\hat{q}_{t+1}(f)$ at finite number of \emph{measurement points} $\bX_\inbatch$ sampled from a distribution $c(\bx)$, %
    i.e., we update $\gamma$ using the gradient of the KL divergence as shown below, where $\eta$ is the learning rate:
    \begin{align} \label{eq:step-kl}
    \left.\gamma_{t+1} = \gamma_t - \eta\nabla_{\gamma}\KL{q_{\gamma}(\bbf_\inbatch)}{\hat{q}_{t+1}(\bbf_\inbatch)}\right|_{\gamma = \gamma_t}.
    \end{align}
    When the likelihood is non-Gaussian, $\hat{q}_{t+1}(\bbf_\inbatch)$ does not have a closed-form expression.
    We propose to upper bound $\KL{q_\gamma (\bbf_\inbatch)}{\hat{q}_{t+1}(\bbf_\inbatch)}$ with the KL divergence between the two joint distributions $\KL{q_\gamma (\bbf_\inbatch, f_n)}{\hat{q}_{t+1}(\bbf_\inbatch,f_n)}$.
    Minimizing this is equivalent to maximizing: %
    \begin{align*}
    &\mathcal{L}_{t}(q_\gamma; q_{\gamma_t}, \bX_\inbatch) = \mathbb{E}_{q_\gamma (\bbf_\inbatch,f_n)}\left[N\beta_t\!\log p(y_n|f_n) +\right. \notag \\ &\left.\beta_t\log p(\bbf_\inbatch,\! f_n)\!+\!(1\!-\!\beta_t)\log q_{\gamma_t}\!(\bbf_\inbatch,\!f_n) \!-\!\log q_\gamma\!(\bbf_\inbatch,\!f_n)\right].
    \end{align*}
    Our method is similar in spirit to RL methods such as 
    temporal difference learning with function approximators~\citep{sutton1998rl}, which also employs stochastic gradients to bootstrap from existing value function approximation. Their success indicates that 
    taking a gradient step with a small step size here might ensure good performance in practice.

    \begin{algorithm}[t]
        \caption{GPNet for supervised learning}
        \label{alg:gp-bnn}
        \begin{algorithmic}[1]
            \REQUIRE $\{(\bx_n, y_n)\}_{n=1}^N$, $c(\bx)$, $M$, $T$, $\beta$, $\eta$.
            \STATE Initialize the inference network $q_{\gamma}$.
            \FOR { $t = 1, \dots, T$ }
            \STATE Randomly sample a training data $(\bx_n, y_n)$.
            \STATE Sample $\bX_\inbatch = (\bx_1, \dots, \bx_M)$ from $c(\bx)$.  
            \IF{Gaussian likelihood}
            \STATE Compute $\hat{q}_{t+1}(\bbf_\inbatch)$ using \eqref{eq:frac-bayes-approx-gpreg}.
            \STATE
            $\gamma_{t+1} \gets \gamma_t - \eta\nabla_{\gamma}\KL{q_{\gamma}(\bbf_\inbatch)}{\hat{q}_{t+1}(\bbf_\inbatch)}$.
            \ELSE
            \STATE
            $\gamma_{t+1} \gets \gamma_t + \eta\nabla_{\gamma}\mathcal{L}_t( q_{\gamma};  q_{\gamma_t},  \bX_\inbatch)$.
            \ENDIF
            \ENDFOR
            \STATE \textbf{return} $q_{\gamma_t}$.
        \end{algorithmic}
    \end{algorithm}
    
    \subsection{Algorithm}
    \label{sec:algo}
    
    We name our algorithm \emph{Gaussian Process Inference Networks} (GPNet), which is summarized in \Cref{alg:gp-bnn}. %
    For each iteration of our algorithm, the stochastic mirror descent update is computed by subsampling a datapoint from the training set, then the inference network is trained to track the update at a set of measurement locations sampled from $c(\bx)$. Though we have described the algorithm using a single data example, it is straightforward to extend it to minibatches.%
    
    The computation cost in case of GP regression is the cost of matrix inversion which is $\bigO(M^3)$. Note that, unlike sparse methods, $M$ does not have to be large for a flexible inference. The inference network can be a complex model containing neural networks which can be very flexible. Our procedure essentially uses $M$ locations to be able to compute stochastic gradients to update the parameters of the network. 

    \vspace{-0.2cm}
    \paragraph{Choice of $c(\bx)$:} Previous works on function-space inference~\citep{sun2018functional,wang2018function,hafner2018reliable} have studied ways to sample the measurement points. The general approach is to apply uniform sampling in the input domain for low-dimensional problems; while for high-dimensional input space, we can sample "near" the training data by adding noise to them. A useful trick for RBF kernels is to set $c(\bx)$ as the training distribution convolved with the kernel. In applications where the input region of test points is known, we can set the $c(\bx)$ to include it. %
    \vspace{-0.3cm}
    \paragraph{Hyperparameter selection:} We set $\beta_t = \beta_0(1 + \xi\sqrt{t})^{-1}$ to ensure that the original stochastic mirror descent converges. Typical values are $\{1, 0.1, 0.01\}$ for $\beta_0$, and $\{1, 0.1\}$ for $\xi$. We can %
    update GP hyperparameters when needed in an online fashion using the lower bound of minibatch log marginal likelihood: $\sum_{i\in B}\mathbb{E}_{q_t} \sum_{i\in B}\log p(y_i|f_i) - \KL{q_t(f_i)}{p(f_i)}$, which is similar to sparse variational methods. In our experiments the learning rate for GP hyperparameters is the same as $\eta$. %
    
    \section{Examples of Inference Networks for GPs}
    \label{sec:infer-net}
    
    So far we have used $q_{\gamma}(f)$ to denote the inference networks, without discussing how to construct them. %
    Below we explore several types of networks that can be used in GPNet.
    
    \vspace{-0.1cm}
    \subsection{Bayesian Neural Networks (BNN)}
    
    A well-known fact in the community is: The function defined by a single-layer fully-connected neural network (NN) with infinitely many hidden units and independent weight randomness is equivalent to a GP~\citep{neal1995bayesian}. Recently, the result is extended to deep NNs~\citep{lee2018deep,matthews2018gaussian,garriga-alonso2018deep,novak2019bayesian}. This equivalence has motivated the use of BNNs as inference networks to model the distribution of functions~\citep{sun2018functional,wang2018function}. Given a neural network $g(\bx; \bOmega)$, where $\Omega$ denotes the network weights, a BNN is constructed by introducing weight randomness: $\bOmega \sim \mathcal{N}(\bOmega_0, \bV)$. %
    Typically %
    $\bV$ is  a factorized or matrix-variate Gaussian that factorizes across layers. In this case the inference network parameters should be $\gamma = \{\bOmega_0, \bV\}$.
    
    However, this approach has several problems. First, the output density of BNN is intractable. %
    In \citet{flam2017mapping,wang2018function} this difficulty is addressed by approximating the output distribution as a Gaussian and estimating the moments from samples, but drawing samples is costly due to many forward passes through the NN. The covariance estimate will present large variance for the typical sample size we can afford. %
    In \citet{sun2018functional} the situation is improved by directly estimating the gradients $\nabla_{\bbf_\inbatch}\log q(\bbf_\inbatch)$ instead of the moments, with a low-variance but biased gradient estimator~\citep{shi2018spectral}. However, the sample size still needs to be hundreds. Moreover, because the stochastic process defined by finite-width BNNs may not be a GP, it is unclear whether matching finite marginal distributions would suffice to match two processes.
    
    \subsection{Tractable Variants}

    Given the problems faced with BNNs, we explore two more types of inference networks that are equivalent to GPs. %
    Both approaches naturally arise from the feature-space representation of GPs. It is known that for a Bayesian linear regression with input feature $\bphi(\bx)$ and Gaussian weights $\bw\sim \mathcal{N}(\bzero, \bSigma)$, the output distribution is equivalent to a GP with the covariance $k(\bx,\bx') = \bphi(\bx)^\top\bSigma\bphi(\bx')$~\citep{rasmussen2006gaussian}. In general, any positive definite kernel $k(\bx, \bx')$ can be written as the inner product of two feature maps $\bphi(\bx)$ and $\bphi(\bx')$. As long as we know the $\bphi(\bx)$ that corresponds to the kernel, we can interpret our GP latent function $f\sim \mathcal{GP}(0, k)$ as a parametric model:
    \begin{equation} \label{eq:feature-gp}
    f(\bx) = \bw^\top \bphi(\bx),\quad\bw \sim \mathcal{N}(\bzero, \bI).
    \end{equation}
    We could define the variational process in a similar form:
    \begin{equation} \label{eq:feature-var-gp}
    q(f): f(\bx) = \bw^\top \bphi(\bx)
    ,\quad \bw \sim \mathcal{N}(\bbm, \bV),
    \end{equation}
    where $\{\bbm, \bV\}$ are the parameters of the inference network.

    \vspace{-0.1cm}
    \paragraph{Random Feature Expansion (RFE)} For GPs with stationary kernels (i.e., kernels that only depend on the difference between inputs, $k(\bx, \bx') = \psi(\bx - \bx')$), Bochner's theorem guarantees that the covariance function can be written as a Fourier transform:
    \begin{equation} \label{eq:bochner}
    k(\bx, \bx') = \int e^{i\bs^\top(\bx - \bx')} p(\bs)d\bs,
    \end{equation}
    where $p(\bs)$ is a spectral density in one-to-one correspondence with $\psi$. Random Fourier features~\citep{rahimi2008random} is an approximation to kernel methods which gives explicit feature maps. The key observation is that \cref{eq:bochner} can be approximated by Monte-Carlo:
    \begin{equation*}
    k(\bx, \bx') \approx \frac{1}{M}\sum_{m=1}^M \cos(\bs_m^\top(\bx - \bx')),\quad\bs_{1:M} \sim p(\bs),
    \end{equation*}
    where the imaginary part is zero. Defining $\bphi_{r}(\bx) = \frac{1}{\sqrt{M}}[\cos(\bs_1^\top\bx), \dots, \cos(\bs_M^\top\bx), \sin(\bs_1^\top\bx), \dots, \sin(\bs_M^\top\bx)]^\top$, we can use it as the approximate feature map: $k(\bx, \bx') \approx \bphi_{r}(\bx)^\top\bphi_{r}(\bx')$. When using random Fourier features, the inference network $\bw^\top\bphi_r(\bx)$ is a neural network with one hidden layer. The activation functions for the hidden layer are $\cos$ and $\sin$. $\bs_1, \dots, \bs_M$ and $\bw$ serve as the input-to-hidden and the hidden-to-output weights, respectively. This architecture is called \emph{Random Feature Expansion} in \citet{cutajar2016random}, where they use a multi-layer stack to mimic a deep GP prior, though inference is still in the weight space. As done in their work, we relax $\bs_1, \dots, \bs_M$ to be trainable so that the inference network parameters are $\gamma = \{\bbm, \bV, \bs_{1:M}\}$. 
    
    \vspace{-0.2cm}
    \paragraph{Deep Neural Networks}
    It is not always easy to explicitly write the inner-product form of a given kernel except the stationary case discussed. We may need a black-box approach, e.g., to parameterize $\bphi(\bx)$ with a function approximator such as neural networks with general nonlinearities~(e.g., ReLU and $\tanh$) and fit it during inference.
    In~\citet{snoek2015scalable} a similar architecture called \emph{adaptive basis regression} networks was proposed as priors in Bayesian optimization.
    Similar to their observations, we have found that this type of networks require significant effort to tune.
    Inspired by the Fisher kernel~\citep{jaakkola1999exploiting}, we also considered an alternative feature map: the vector of how much information stored in each weight parameter, measured by the gradients of network outputs with respect to weights $\nabla_{\bOmega}g(\bx;\bOmega)$. 
    Plugging into $\bphi(\bx)$, we get a \emph{Neural Tangent Kernel}~\citep{NIPS2018_8076}:
    $
    k_{\text{NTK}}(\bx, \bx') = \nabla_{\bOmega}g(\bx;\bOmega_0)^\top\bV\nabla_{\bOmega}g(\bx;\bOmega_0).$
    The performance of an inference network parameterized in this way is similar to a BNN because the NTK can be interpreted as introducing weight randomness on a first-order expansion of neural networks:
    \begin{align*}
    g(\bx; \bOmega) = g(\bx; \bOmega_0) + \nabla_{\bOmega}g(\bx;\bOmega_0)(\bOmega - \bOmega_0),
    \end{align*}
    where if $\bOmega \sim \mathcal{N}(\bOmega_0, \bV)$ then it is equivalent to the GP: $f\sim \mathcal{GP}(g(\cdot; \bOmega_0), k_{\text{NTK}})$.
    As seen, the mean function is as flexible as a deep NN, while the kernel utilizes the gradients as features. %
    In some of our experiments we found it faster to converge than adaptive-basis ones, but the drawback is much higher computational cost due to backpropagation through Jacobians.

    \begin{figure*}[t]\vspace{-0.1cm}
        \begin{subfigure}[b]{0.68\linewidth}
            \includegraphics[height=2.5cm]{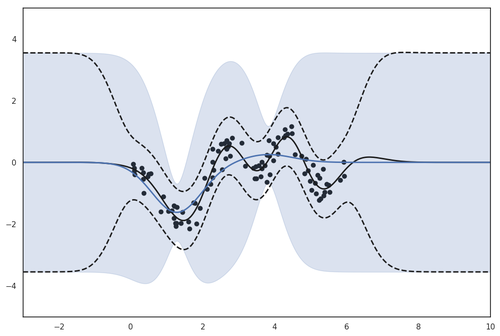}
            \includegraphics[height=2.5cm]{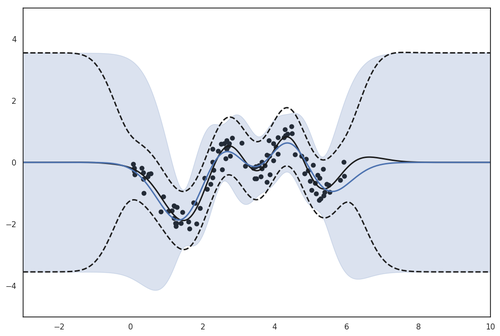}
            \includegraphics[height=2.5cm]{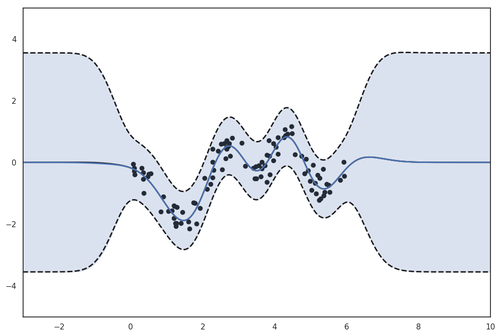} \\
            \includegraphics[height=2.5cm]{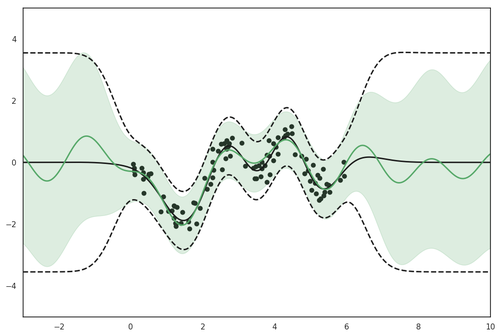}
            \includegraphics[height=2.5cm]{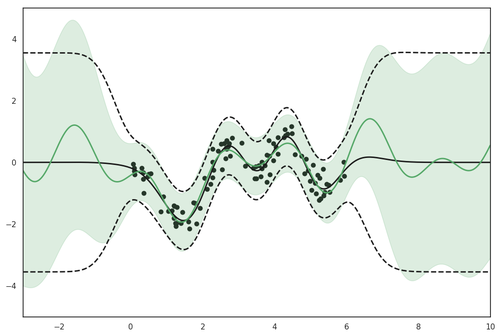}
            \includegraphics[height=2.5cm]{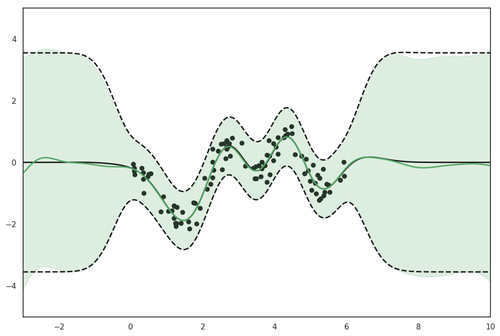}
            \caption{}
            \label{fig:toy-vs-svgp}
        \end{subfigure}
        \begin{subfigure}[b]{0.3\linewidth}
            \centering       
            \includegraphics[height=2.5cm]{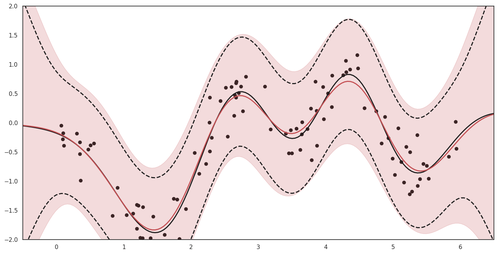} \\
            \centering
            \includegraphics[height=2.5cm]{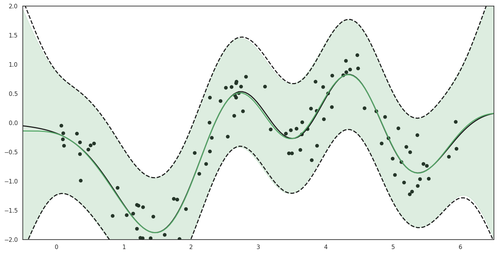}
            \caption{}
            \label{fig:toy-vs-fbnn}
        \end{subfigure}
        \vspace{-0.1cm}
        \caption{Posterior process on the Snelson dataset, where shaded areas correspond to intervals of $\pm3$ standard deviations, and dashed lines denote the ground truths. (a) \textbf{Top row}: SVGP with $M\in\{2, 5, 20\}$ (left to right) inducing points; \textbf{Bottom row}: GPNet with $M\in\{2, 5, 20\}$ measurement points. (b) \textbf{Top}: FBNN, $M=20$; \textbf{Bottom}: GPNet, $M=20$.}
        \vspace{-0.1cm}
        \label{fig:snelson}
    \end{figure*}

    \section{Experiments}
    
    Throughout all experiments, $M$ denotes both the number of inducing points in SVGP and the number of measurement points in GPNet and FBNN~\citep{sun2018functional}.
    Implementations are based on a customized version of GPflow~\citep{GPflow2017,sun2018differentiable} and ZhuSuan~\citep{zhusuan2017}. Code is available at \url{https://github.com/thjashin/gp-infer-net}.
    
    \subsection{Synthetic Data}
    
    We consider the inference of a GP with RBF kernel on the synthetic dataset introduced in \citet{snelson2006sparse}. We analyze the properties of our method and compare with SVGP and FBNN~\citep{sun2018functional}. %
    We fit these algorithms with minibatch size 20 on 100 data points. We ran for 40K iterations and used learning rate 0.003 for all methods. For fair comparison, for all three methods we pretrain the prior hyperparameters for 100 iterations using the GP marginal likelihood and keep them fixed thereafter. We vary $M$ in $\{2, 5, 20\}$ for all methods. The networks used in GPNet and FBNN are the same RFE with 20 hidden units.
    
    Results are plotted in Fig.~\ref{fig:snelson}. We can see that the performance of SVGP grows with more inducing points. %
    When $M=20$, both SVGP and GPNet can recover the exact GP prediction. %
    GPNet fits the data better when $M=2,5$. This is because $M$ does not constrain the capacity of the inference network, though it does affect the convergence speed, i.e., smaller $M$ causes larger variance in the training.
    In Fig.~\ref{fig:toy-vs-fbnn} we take a closer look at the predictions by GPNet and FBNN near the training data. %
    We can see that FBNN consistently overestimates the uncertainty. This effect can be well explained by their heuristic way of doing minibatch, that in each iteration they fit a different objective to match the local effect of 20 training points in a minibatch, while our stochastic mirror descent maintains a shared global objective that takes all observations into consideration.

    \begin{figure*}[t]%
        \centering
        \begin{subfigure}[t]{0.20\textwidth}
            \centering
            \includegraphics[height=3cm]{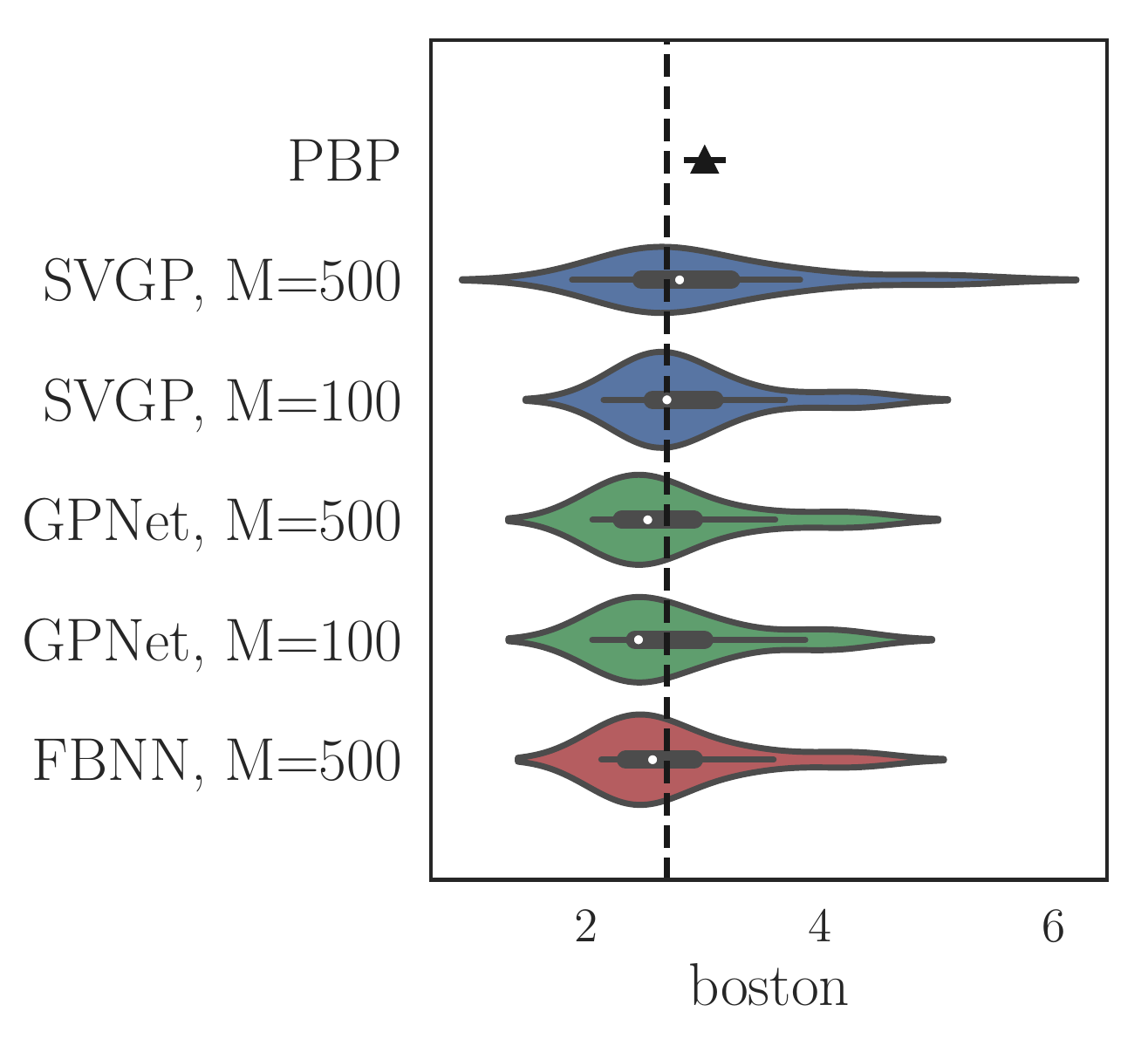}
        \end{subfigure}
        \begin{subfigure}[t]{0.13\textwidth}
            \centering
            \includegraphics[height=3cm]{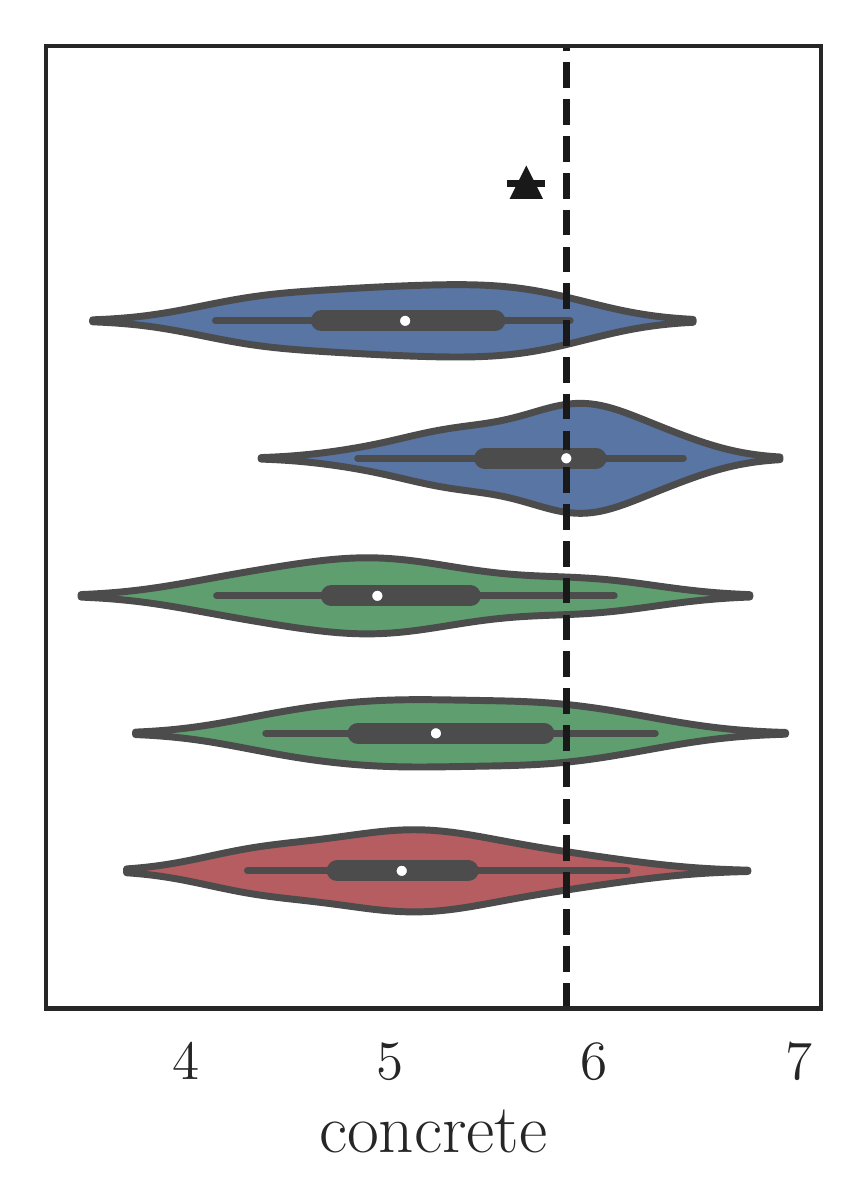}
        \end{subfigure}
        \begin{subfigure}[t]{0.13\textwidth}
            \centering
            \includegraphics[height=3cm]{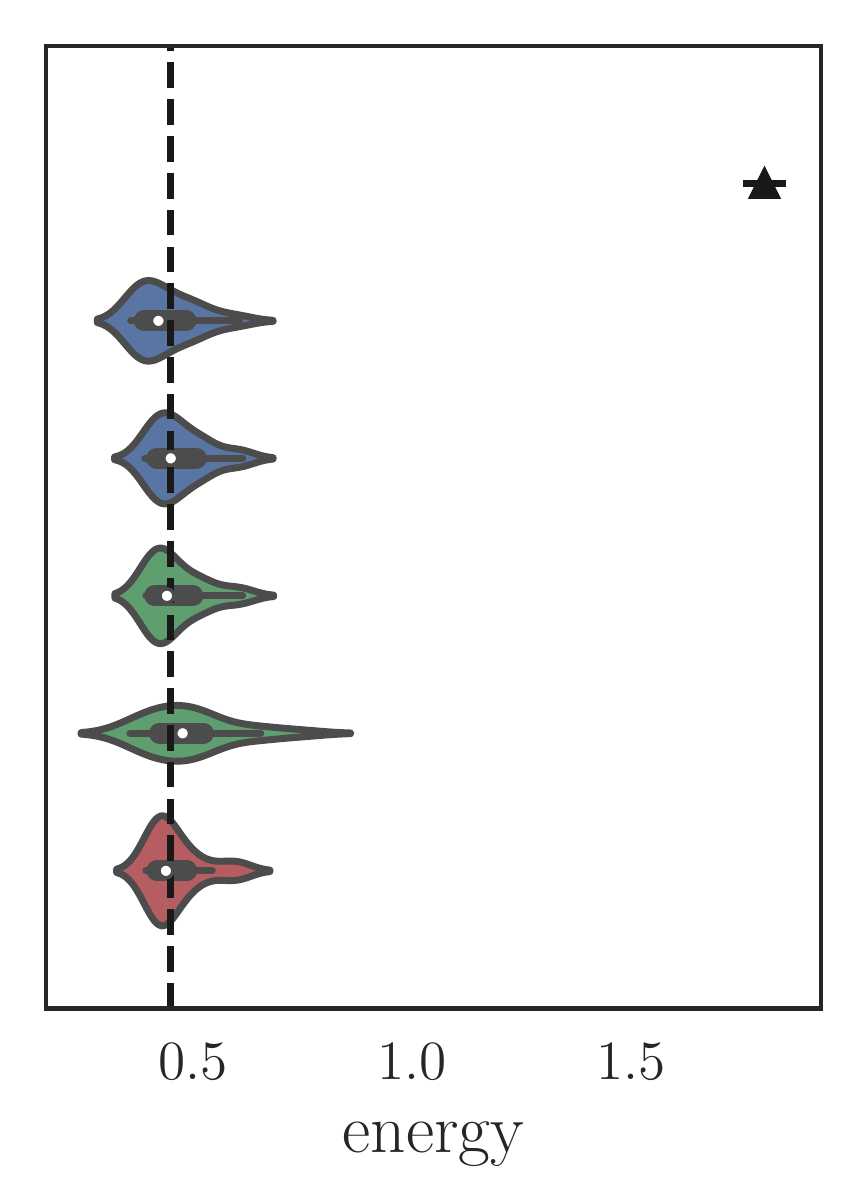}
        \end{subfigure}
        \begin{subfigure}[t]{0.13\textwidth}
            \centering
            \includegraphics[height=3cm]{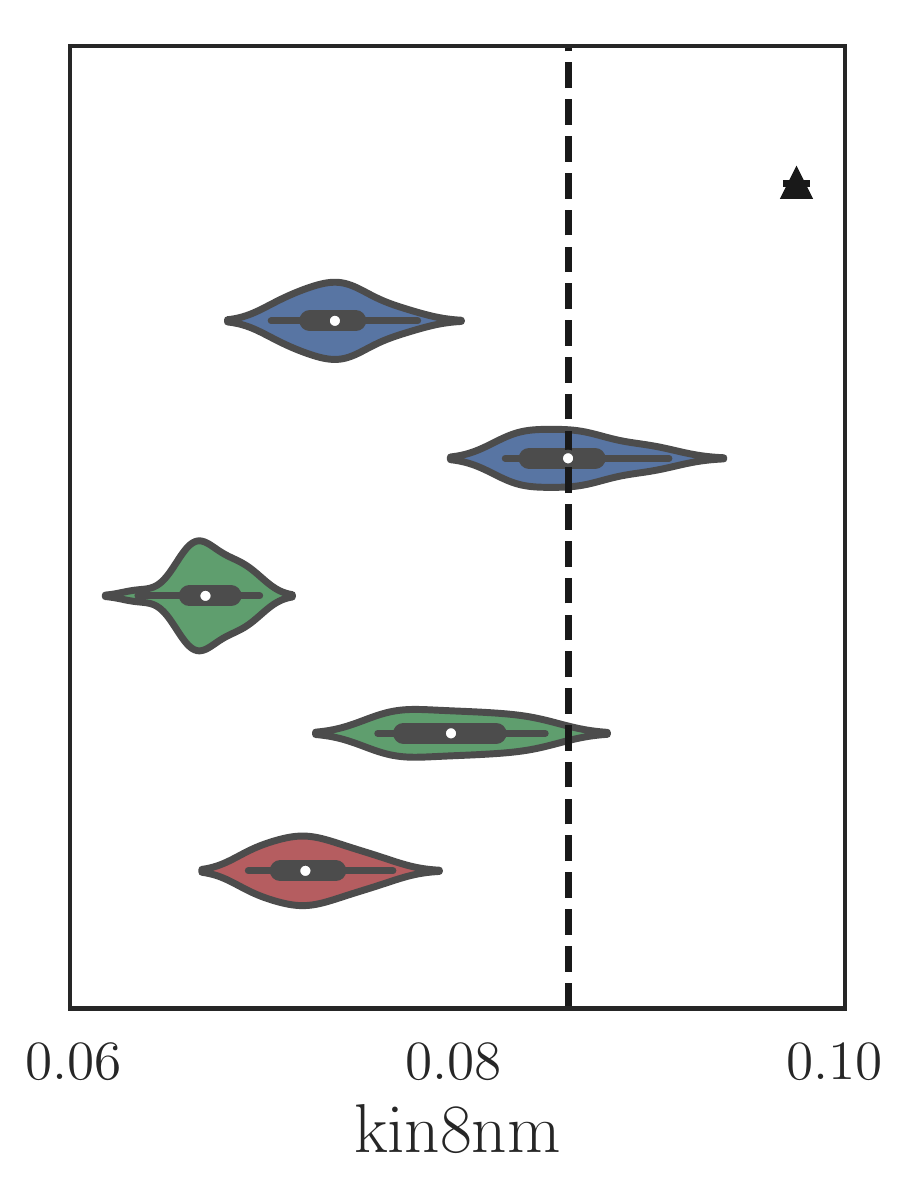}
        \end{subfigure}
        \begin{subfigure}[t]{0.13\textwidth}
            \centering
            \includegraphics[height=3cm]{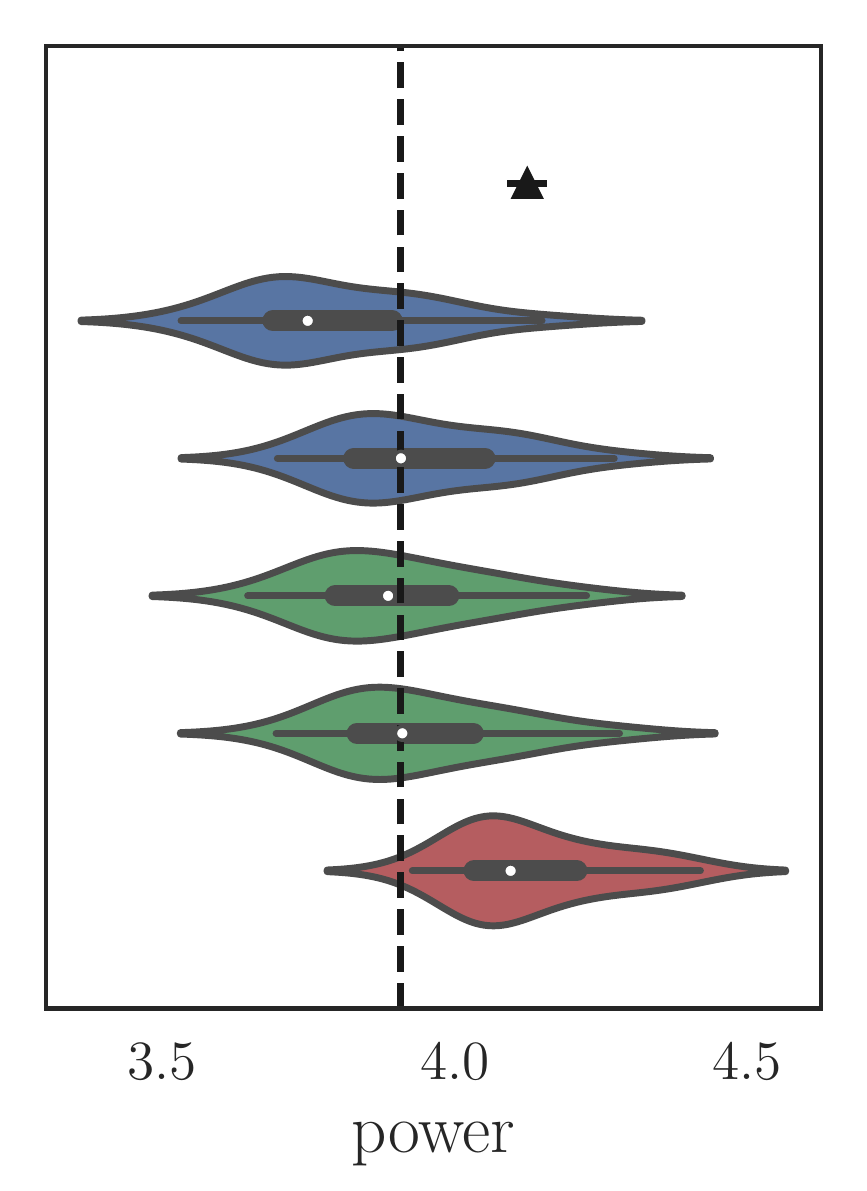}
        \end{subfigure}
        \begin{subfigure}[t]{0.13\textwidth}
            \centering
            \includegraphics[height=3cm]{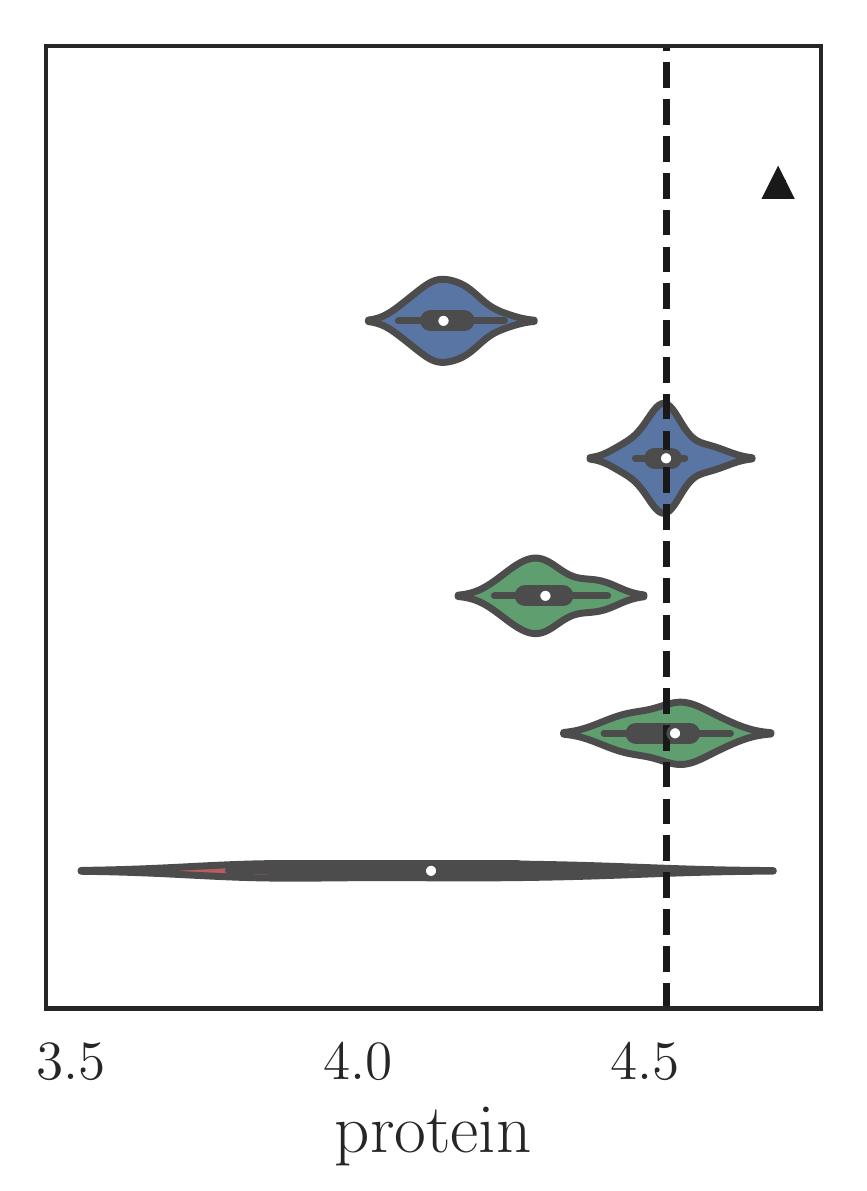}
        \end{subfigure}
        \begin{subfigure}[t]{0.12\textwidth}
            \centering
            \includegraphics[height=3cm]{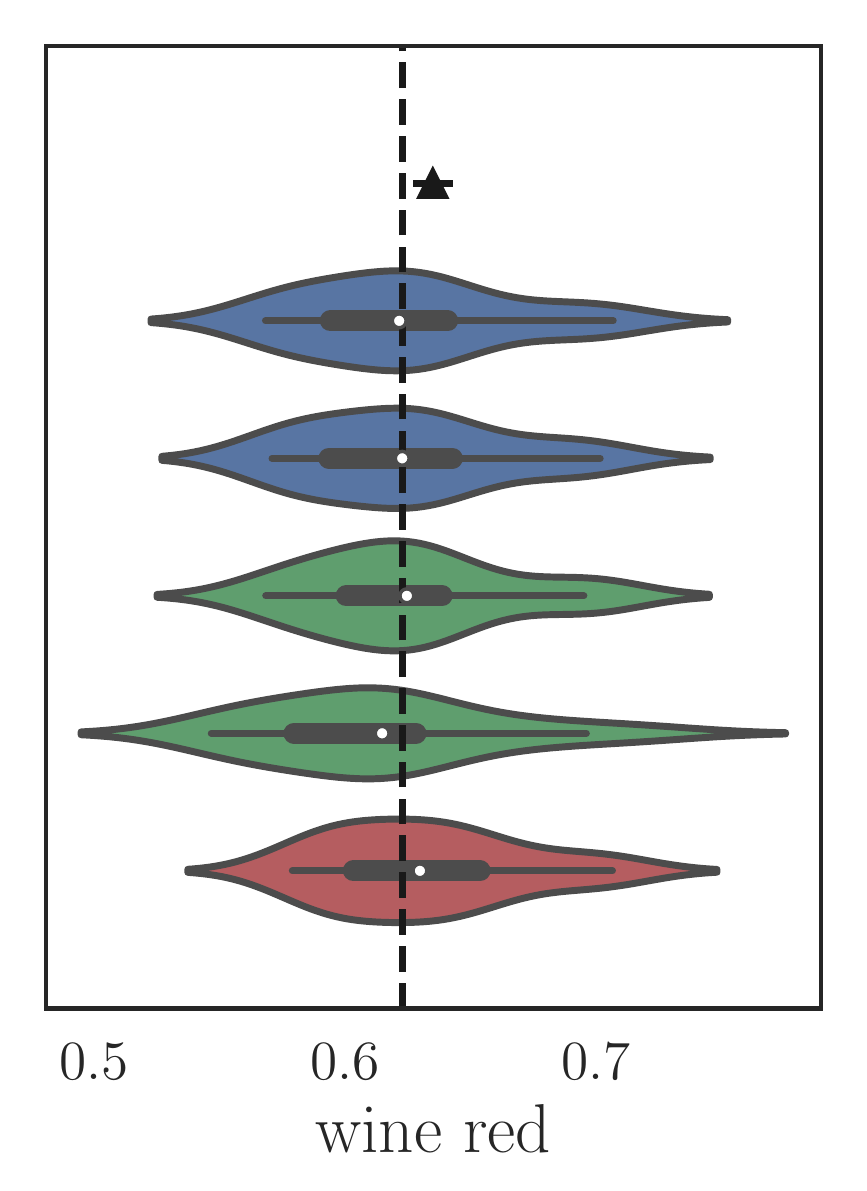}
        \end{subfigure} \\
        \centering
        \begin{subfigure}[t]{0.20\textwidth}
            \centering
            \includegraphics[height=3cm]{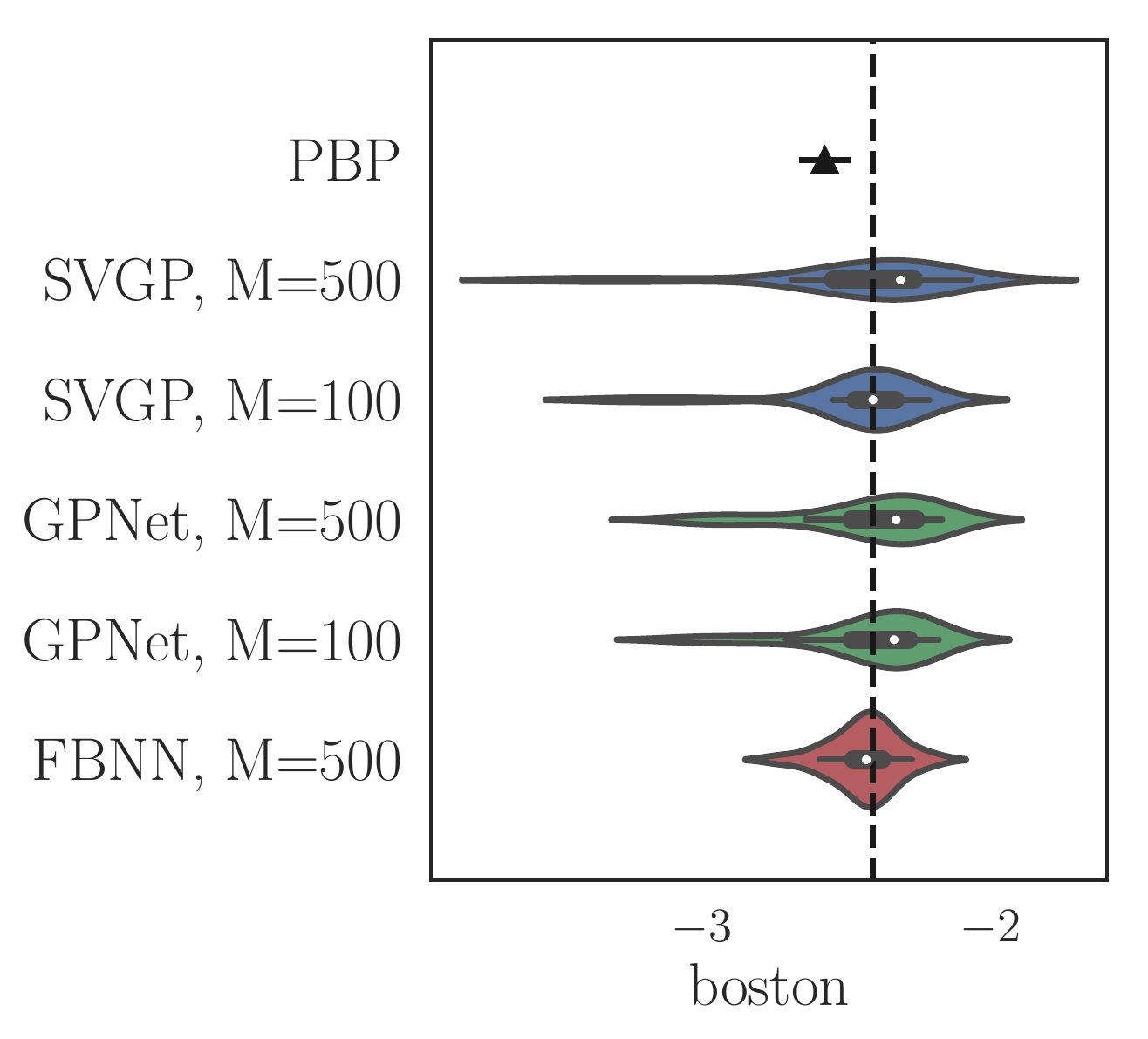}
        \end{subfigure}
        \begin{subfigure}[t]{0.13\textwidth}
            \centering
            \includegraphics[height=3cm]{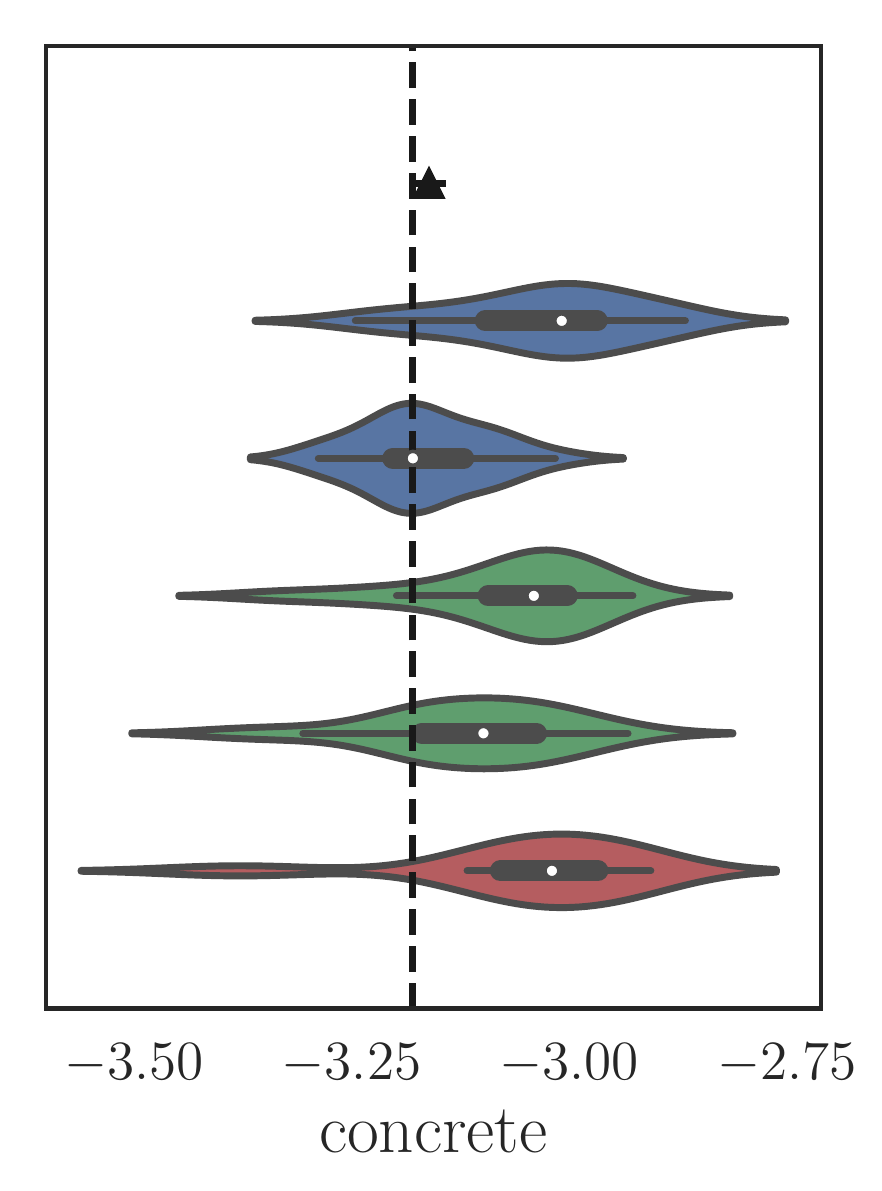}
        \end{subfigure}
        \begin{subfigure}[t]{0.13\textwidth}
            \centering
            \includegraphics[height=3cm]{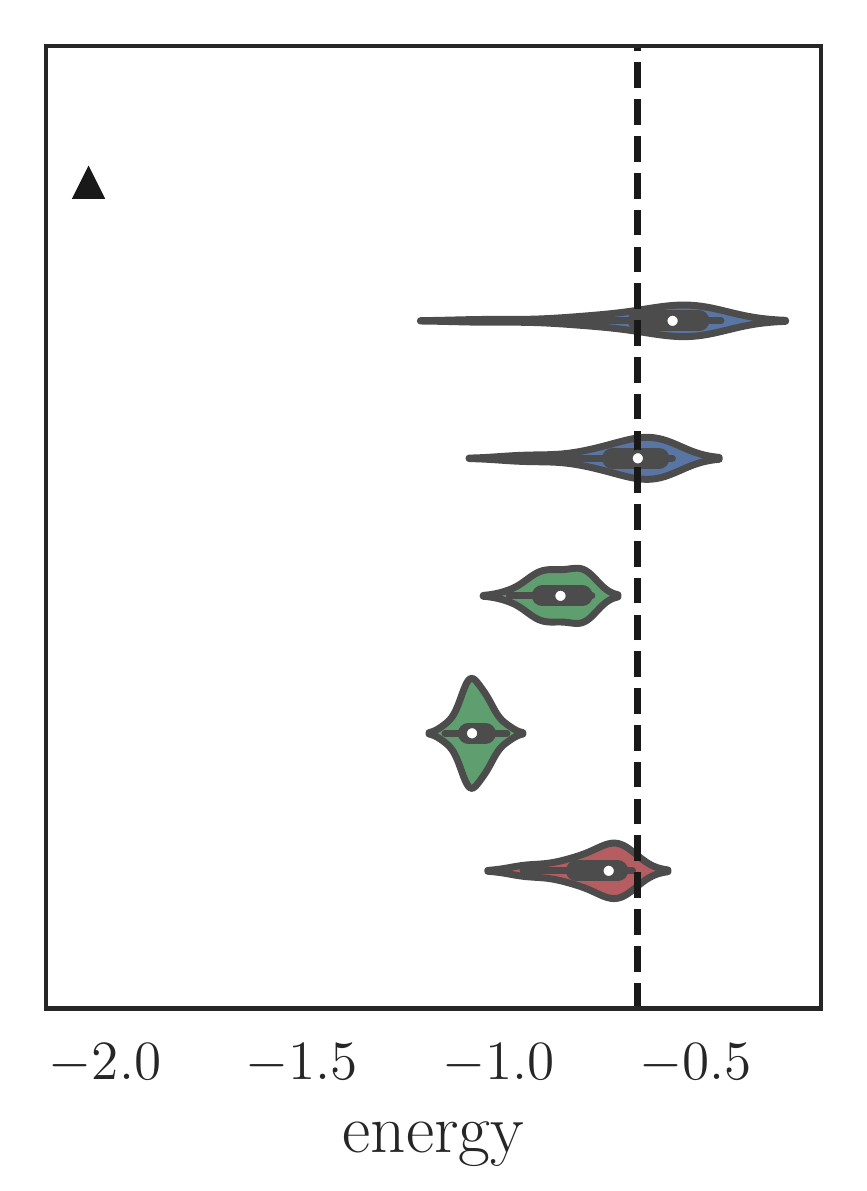}
        \end{subfigure}
        \begin{subfigure}[t]{0.13\textwidth}
            \centering
            \includegraphics[height=3cm]{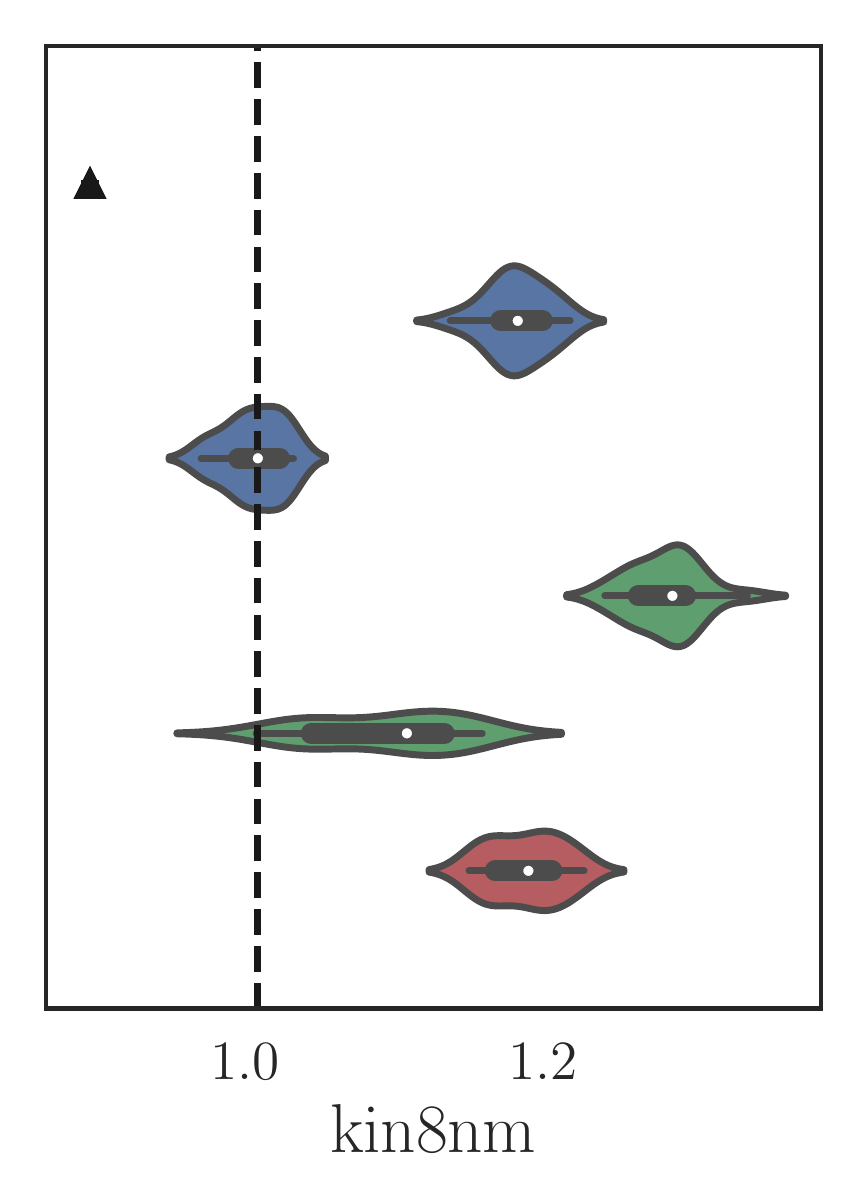}
        \end{subfigure}
        \begin{subfigure}[t]{0.13\textwidth}
            \centering
            \includegraphics[height=3cm]{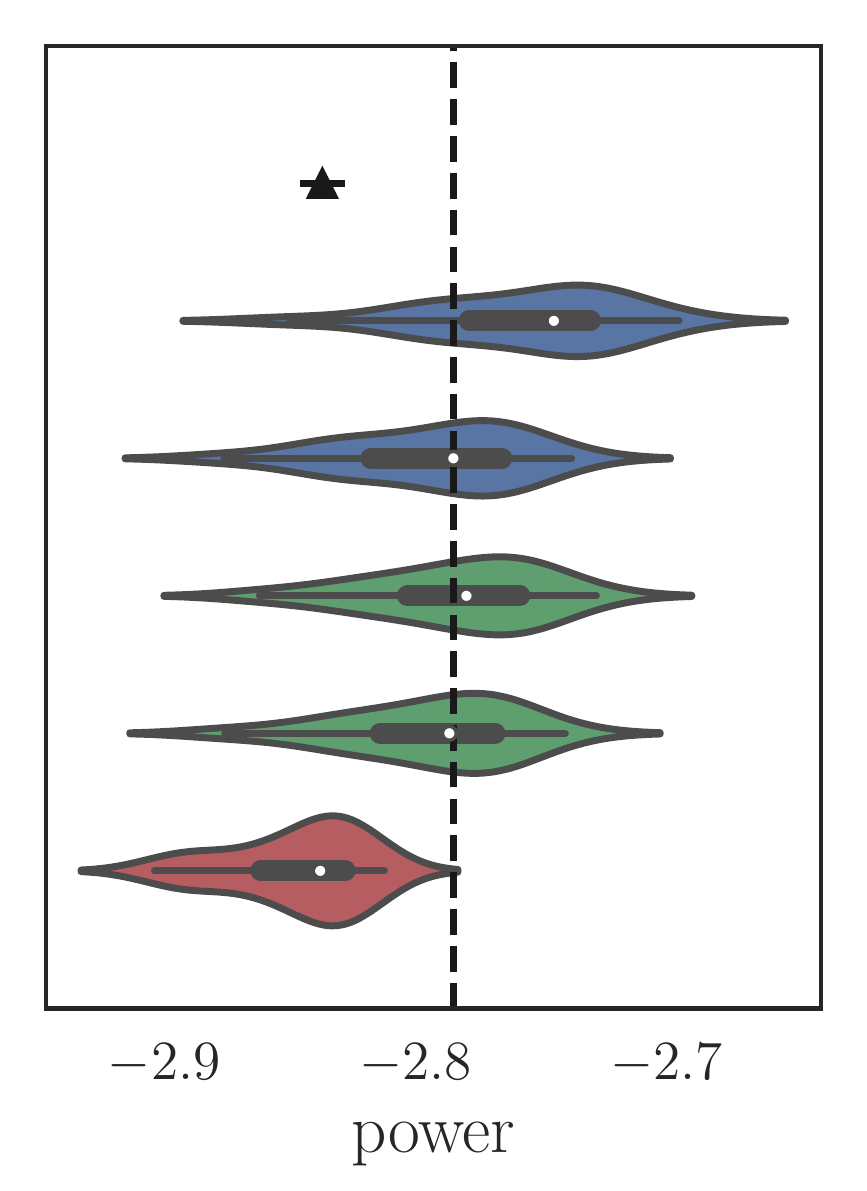}
        \end{subfigure}
        \begin{subfigure}[t]{0.13\textwidth}
            \centering
            \includegraphics[height=3cm]{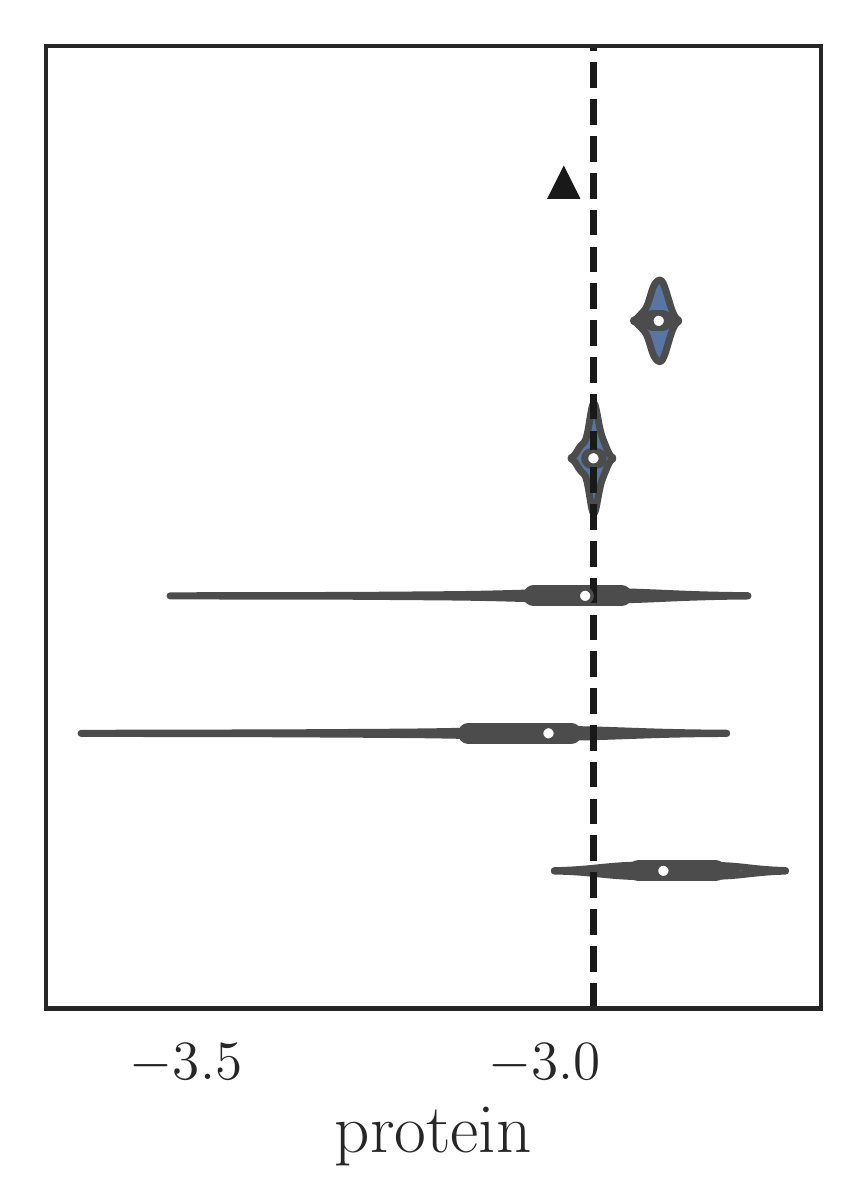}
        \end{subfigure}
        \begin{subfigure}[t]{0.12\textwidth}
            \centering
            \includegraphics[height=3cm]{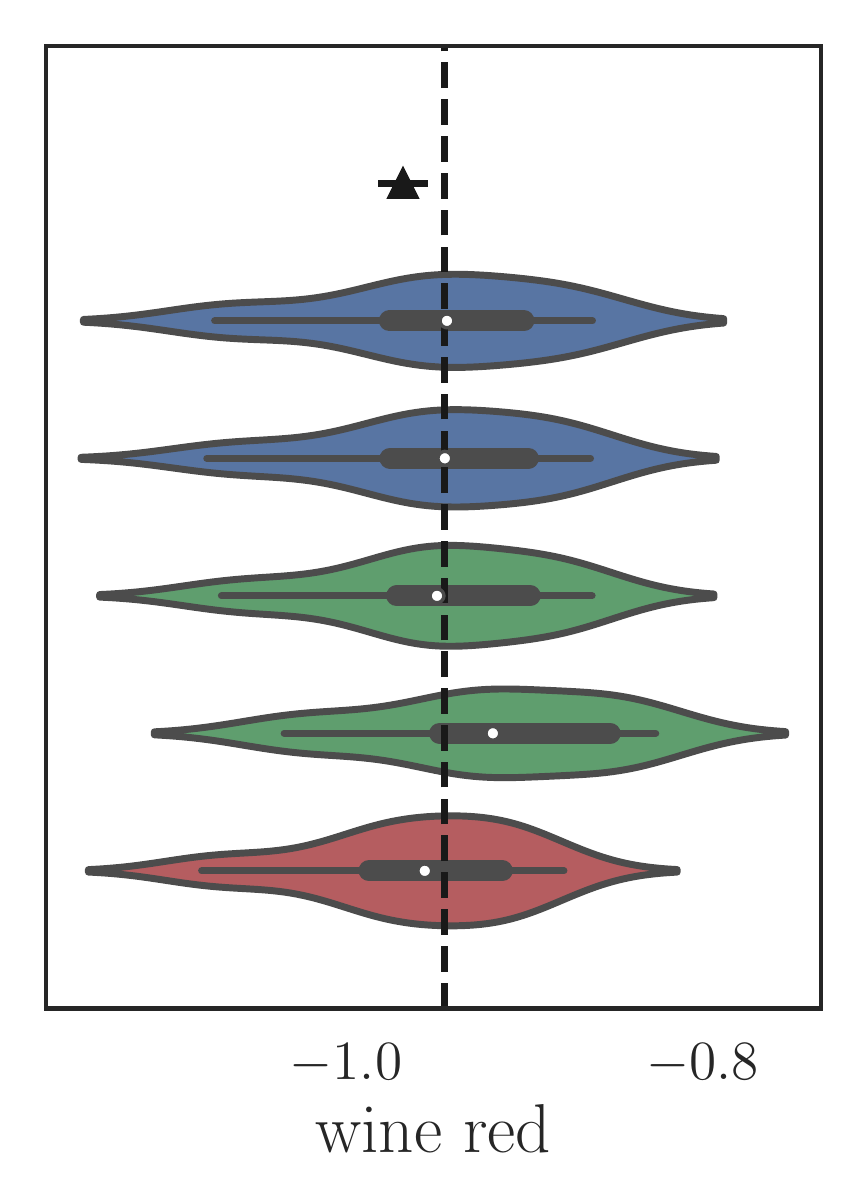}
        \end{subfigure}
        \caption{Regression results on benchmark datasets. \textbf{Top row}: Test RMSE; \textbf{Bottom row}: Test log likelihood. All results are drawn as violin plots which shows a kernel density of results on different splits, except for PBP we only have the mean and standard error, so an error bar is drawn instead.}
        \label{fig:regression}
    \end{figure*}
    
    \subsection{Regression}
    \vspace{0.05cm}
    
    \paragraph{Benchmarks}
    We evaluate our method on seven standard regression benchmark datasets. %
    We use RFE networks with 1000 hidden units for this task.
    Following the settings of \citet{salimbeni2017doubly}, we compare to the strong baselines: SVGP with 100 and 500 inducing points.
    We also compare to FBNN using the same inference network as in GPNet\footnote{Though the algorithm of \citet{sun2018functional} is designed for BNNs, it also applies to other types of inference networks.}.
    To put the comparison into a wider context, we include the results by probabilistic backpropagation~(PBP)~\citep{hernandez2015probabilistic}, which is an effective weight-space inference method for BNNs. Details of datasets and experiment settings can be found in \cref{app:exp-details}.
    Results are summarized in Fig.~\ref{fig:regression}. %
    We can see that GPNet has comparable or smaller RMSE than SVGP on most datasets, and the performance gap is often large when comparing them given $M=100$.
    This demonstrates the effectiveness of inference networks than inducing points given similar computational complexity.
    The regression results on small datasets including Boston, Concrete and Energy show that overfitting is not observed with our powerful networks. 
    Note that these three datasets only contain 1-2 minibatches of data, thus the performance of FBNN and GPNet are comparable because minibatch training is not an issue; while on larger datasets such as Kin8nm, Power and Wine, GPNet consistently outperforms FBNN. We find on Protein GPNet is slower to converge than other methods, and increasing the training iterations will give far better performance.
    
    \begin{table}[t]\vspace{-0.4cm}
        \caption{Large-scale regression on the airline dataset.}
        \label{tab:airline}
        \begin{center}
            \resizebox{\columnwidth}{!}{%
                \begin{small}
                    \begin{tabular}{l|ccc|ccc}
                        \toprule
                        \multirow{2}{*}{\textsc{Metric}} & \multicolumn{3}{c|}{M=100} & \multicolumn{3}{c}{M=500} \\
                        \cmidrule{2-4}  \cmidrule{5-7}
                        & \textsc{SVGP} & \textsc{GPNet} & \textsc{FBNN} & \textsc{SVGP} & \textsc{GPNet} & \textsc{FBNN}\\
                        \midrule
                        RMSE & 24.261 & 24.055 & 23.801 & 23.698 & \bf 23.675 & 24.114 \\
                        Test LL & -4.618 & -4.616 & -4.586 & -4.594 & -4.601 & \bf -4.582 \\
                        \bottomrule
                    \end{tabular}
            \end{small}}
        \end{center}
        \vspace{-0.6cm}
    \end{table}

    \vspace{-0.2cm}
    \paragraph{Airline Delay} To demonstrate the advantage of our minibatch algorithm on large-scale datasets, we conducted experiments on the airline delay dataset, which includes 5.9 million flight records in the USA from Jan to Apr in 2018. Following the protocol in~\citet{hensman2013gaussian}, we randomly take 700K points for training and 100K for testing. The results are shown in \Cref{tab:airline}. Experiment details can be found in \cref{app:exp-details}. We can see that GPNet achieves best RMSE among three methods and has comparable test log likelihoods with SVGP. The RMSE gap between $M=100$ and $M=500$ for SVGP is larger than that of GPNet, which again demonstrates that the power of our inference network is not limited by $M$. Interestingly, larger $M$ seems to cause underfitting of FBNN and leads to worse RMSE, which may also be due to the minibatch issue.
    
    \subsection{Classification}
    
    Finally, we demonstrate the flexibility of GPNet by fitting a deep convolutional inference network for a CNN-GP~\citep{garriga-alonso2018deep}, whose covariance kernel are derived from an infinite-width Bayesian ConvNet (see \cref{app:exp-cls} for detailed derivation of the GP prior).
    
    \begin{table}[t]\vspace{-0.2cm}
        \caption{Image classification: Test error rates.}
        \label{tab:classification}
        \begin{center}
            \begin{small}
                \resizebox{\columnwidth}{!}{%
                    \begin{tabular}{lcc}
                        \toprule
                        \textsc{Methods} & \textsc{MNIST} & \textsc{CIFAR10} \\
                        \midrule
                        SVGP, RBF-ARD~{\scriptsize\citep{krauth2016autogp}} & 1.55\% & - \\
                        Conv GP~{\scriptsize\citep{van2017convolutional}} & 1.22\% & 35.4\% \\
                        SVGP, CNN-GP~{\scriptsize\citep{garriga-alonso2018deep}} & 2.4\% & - \\
                        GPNet, CNN-GP & 1.12\% &  \bf 24.63\% \\
                        \midrule
                        NN-GP~{\scriptsize\citep{lee2018deep}} & 1.21\% & 44.34\% \\
                        CNN-GP~{\scriptsize\citep{garriga-alonso2018deep}} & 0.96\% & - \\
                        ResNet-GP~{\scriptsize\citep{garriga-alonso2018deep}} & \bf 0.84\% & - \\
                        CNN-GP~{\scriptsize\citep{novak2019bayesian}} & 0.88\% & 32.86\% \\
                        \bottomrule
                    \end{tabular}%
                }
            \end{small}
        \end{center}
        \vspace{-0.6cm}
    \end{table}
    
    Previously, inference for GPs with such a kernel has only been investigated through exact prediction~\citep{garriga-alonso2018deep,novak2019bayesian}, where the classification problem is treated as regression so that \cref{eq:gpreg} applies. Though this is done for datasets like MNIST and CIFAR10 in recent works, $O(N^3)$ complexity is impractical for the method to be widely adopted. A scalable option would be %
    sparse approximations. We tried SVGP for this prior. However, we found the training is unstable if we update the inducing point locations. Initializing them with data or with K-means centers both result in numerical errors that prevent the method from learning. There are no other results reported using SVGP for such kernels except in \citet{garriga-alonso2018deep}, where they also fix the inducing point locations (1000 training data were used). We believe it is due to the difficulty of finding good inducing point locations in such a high-dimensional input space of images.
    
    We test GPNet on MNIST and CIFAR10 with a CNN-GP prior. The ConvNet that defines this prior has 6 residual blocks (details in \cref{app:exp-details}). It is natural to use the original ConvNet with trainable weight randomness as the inference network ($q(f)$) for this CNN-GP. However, as discussed in \cref{sec:infer-net}, using a BNN results in intractable output distributions which require many efforts to address. %
    To avoid this, we use a deterministic ConvNet with an NTK on top of it defined using the fully-connected layers. This enables flexible covariance modeling while still allowing an efficient training. With the non-conjugate form of our algorithm, we are free to use a softmax likelihood, which is more suitable to classification tasks.
    
    Results are compared to recent works in \cref{tab:classification}. The first half of the table are approximate inference approaches with classification likelihoods, while the second half does exact prediction by GP regression. %
    By comparing to carefully-designed inducing-point approaches such as Conv GP~\cite{van2017convolutional}, we can clearly see the advantage of our method, i.e., easily scaling up GP inference to highly-structured kernels by using flexible inference networks that match the structures, while getting %
    superior performance than carefully-designed inducing-point methods.

    \section{Conclusion}
    We propose an algorithm to scalably train a stochastic inference network to approximate the GP posterior distribution. In the algorithm the inference network is trained by tracking a stochastic functional mirror descent update which is cheap to compute from the current approximation using a minibatch of data. Experiments show that our algorithm fixes the minibatch issue of previous works on function-space inference. Empirical comparisons to sparse variational GP methods show that our method is a more flexible alternative to GP inference.

    \section*{Acknowledgements}
    We thank Ziyu Wang, Shengyang Sun, Ching-An Cheng for helpful discussions and Hugh Salimbeni for help with the experiments.
    JS was supported by a Microsoft Research Asia Fellowship. This work was supported by the National Key
    Research and Development Program of China (No. 2017YFA0700904), NSFC Projects (Nos. 61620106010, 61621136008, 61571261), Beijing NSF Project (No. L172037), DITD Program JCKY2017204B064, Tiangong Institute for Intelligent Computing, NVIDIA NVAIL Program, and the projects from Siemens and Intel.

    \bibliography{example_paper}
    \bibliographystyle{icml2019}
    
    \onecolumn
    \clearpage
    
    \appendix

    \icmltitle{Scalable Training of Inference Networks for Gaussian-Process Models\\Appendix}
    
    \section{Derivation using the Dual Representation of GP and Natural Gradient}
    \label{app:natgrad}
    
    We note that the stochastic mirror descent is equivalent to natural gradient updates for exponential family densities~\citep{raskutti2015information,khan2017conjugate}. 
    To show this, we derive the same adaptive Bayesian filter in \cref{eq:frac-bayes} using the dual representation of GP as a Gaussian measure and natural gradient.
    
        A Gaussian process $\mathcal{GP}(m(\bx), \kappa(\bx, \bx'))$ has a dual representation as a Gaussian measure $\nu$ on a separable Banach space $\mathcal{B}$~\citep{cheng2017variational,mallasto2017learning}.
        There is an RKHS $\mathcal{H}$ that corresponds to a positive definite kernel $k$ densely embedded in $\mathcal{B}$. 
        The measure $\nu$ is constructed as follows. 
        First define a \emph{canonical Gaussian cylinder set measure} $\nu^{\mathcal{H}}$ on $\mathcal{H}$, denoted as $\mathcal{N}(f|\mu, \Sigma)$, where $\mu\in \mathcal{H}$ is the mean function, $\Sigma: \mathcal{H}\to\mathcal{H}$ is a bounded positive semi-definite linear operator. They satisfy
        \begin{align*}
            m(x) &= \mu^\top k(x,\cdot), \\
            \kappa(x, x') &= k(x, \cdot)^\top\Sigma k(x', \cdot),
        \end{align*}
        where $h^\top g$ denotes inner product in the RKHS: $h^\top g = \langle h, g\rangle_\mathcal{H}$.
        Let $i$ be the inclusion map from $\mathcal{H}$ into $\mathcal{B}$.
        Then the measure $\nu$ is induced by $\nu^{\mathcal{H}}$ using this map. 
        In the measure theory of infinite-dimensional space, $\nu$ is known as the \emph{abstract Wiener measure}. 
        The RKHS $\mathcal{H}$ is sometimes called the \emph{Cameron-Martin space}.

    \begin{rem}
        The intuition for this construction is that the canonical Gaussian cylinder set measure $\nu^{\mathcal{H}}$ is not a proper measure. In fact, we can show that countable additivity does not hold for this "measure"~\citep{eldredge2016analysis}. 
        The inclusion map $i$ here radonifies $\nu^{\mathcal{H}}$ into a true measure $\nu$. 
        One way to think about this is that functions drawn from the Gaussian process fall outside of $\mathcal{H}$ with probability one~\citep{kanagawa2018gaussian}, but are contained in $\mathcal{B}$. 
        Despite this, as pointed out in \citet{cheng2017variational}, we can conveniently work with the canonical form $\mathcal{N}(f|\mu, \Sigma)$ and get correct results as long as the conclusion is independent of the dimension of $f$.
    \end{rem}
    
    It is easy to check that the canonical Gaussian measure which corresponds to the GP prior $\mathcal{GP}(0, k(\bx, \bx'))$ is $p(f) = \mathcal{N}(f|0, I)$. 
    Assuming the canonical form of the variational distribution is $q(f) = \mathcal{N}(f|\mu, \Sigma)$, we have stochastic approximation of the lower bound as:
    \begin{equation*}
        \hat{\mathcal{L}}(q) = N\mathbb{E}_{q} \log p(y_n|f) - \KL{q(f)}{p(f)}.
    \end{equation*}
    Now that we can interpret GPs in the form of canonical Gaussian measures, we can then write $q(f)$ and $p(f)$ in the exponential family form to simplify the derivation:
    \begin{align*}
        q(f) &\propto \exp\{\lambda^\top t(f) - A(\lambda) \}, \\
        p(f) &\propto \exp\{\lambda_0^\top t(f) - A(\lambda_0) \}.
    \end{align*}
    where $t(f) = \{f, ff^\top\}$ denotes the sufficient statistics, $A(\lambda) = \frac{1}{2}\mu^\top\Sigma^{-1}\mu + \frac{1}{2}\log|\Sigma|$ is the partition function. The natural parameters of $p(f)$ and $q(f)$ are $\lambda_0 = \{0, -\frac{1}{2}I\}$ and $\lambda = \{\Sigma^{-1}\mu, -\frac{1}{2}\Sigma^{-1}\}$, respectively. Let $u$ denote the mean parameters of $q(f)$: $u = \mathbb{E}_{q}\left[t(f)\right] = \{\mu, \Sigma\}$. There is a dual relationship between the natural parameter $\lambda$ and the mean parameter $u$: $u = \nabla_{\lambda}A(\lambda)$. The mapping $\nabla A$ is one-to-one when the exponential family is minimal~\citep{wainwright2008graphical}. The stochastic natural gradient of the lower bound with respect to $\lambda$ is defined as
    \begin{align*}
        \natgrad_{\lambda} \hat{\mathcal{L}}(q) &= F(\lambda)^{-1} \nabla_{\lambda}\hat{\mathcal{L}}(q),
    \end{align*}
    where $F(\lambda) = \mathbb{E}_{q}\left[\nabla_{\lambda}\log q(f)\; \nabla_{\lambda}\log q(f)^\top\right]$ is the Fisher information matrix. The natural gradient of the KL divergence term is
    \begin{align*}
        \natgrad_{\lambda}\KL{q(f)}{p(f)} &= F(\lambda)^{-1}\nabla_{\lambda}\mathbb{E}_{q}\left[\log \frac{q(f)}{p(f)}\right] \\
        &= F(\lambda)^{-1}\nabla_{\lambda}\left[(\lambda - \lambda_0)^\top u - A(\lambda) + A(\lambda_0)\right] \\
        &= F(\lambda)^{-1}\left[u - \nabla_{\lambda}A(\lambda) + \nabla_{\lambda}u(\lambda - \lambda_0)\right] \\
        &= F(\lambda)^{-1}\nabla^2_{\lambda}A(\lambda)(\lambda - \lambda_0) \\
        &= \lambda - \lambda_0,
    \end{align*}
    where we have used the fact that $\nabla^2_{\lambda} A(\lambda) = F(\lambda)$. We can also derive a simplified form of the natural gradient of the conditional log likelihood term, by writing it as the gradient with respect to the mean parameter $u$:
    \begin{align*}
        \natgrad_{\lambda}\mathbb{E}_q[\log p(y_n|f)] &= F(\lambda)^{-1}\nabla_{\lambda}\mathbb{E}_{q}\left[\log p(y_n|f)\right] \\ 
        &= F(\lambda)^{-1}\nabla_{\lambda}u\cdot\nabla_{u}\mathbb{E}_q[\log p(y_n|f)] \\ 
        &= F(\lambda)^{-1}\nabla^2_{\lambda} A(\lambda)\cdot\nabla_{u}\mathbb{E}_q[\log p(y_n|f)] \\
        &= \nabla_{u}\mathbb{E}_q[\log p(y_n|f)].
    \end{align*}
    So the natural gradient update can be written as
    \begin{align*}
        \lambda_{t+1} &= \lambda_t + \beta_t\left(N\nabla_{u}\mathbb{E}_q\left[\log p(y_n|f)\right] - \lambda_t + \lambda_0\right) \\
        &= (1 - \beta_t)\lambda_t + \beta_t\left(N\nabla_{u}\mathbb{E}_q[\log p(y_n|f)] + \lambda_0\right).
    \end{align*}
    Reinterpreting the above equation in the density space, we have
    \begin{equation} \label{eq:natgrad-vi}
        q_{t+1}(f) \propto q_t(f)^{1 - \beta_t}p(f)^{\beta_t}\exp\{\langle\nabla_{u}\mathbb{E}_q[\log p(y_n|f)], t(f)\rangle\}^{N\beta_t}.
    \end{equation}
    The likelihood $p(y_n|f)$ is said to be conjugate with the prior if it has a form as $p(y_n|f) \propto \exp\{\lambda(y_n)^\top t(f)\}$. For example, in GP regression, the likelihood is $p(y_n|f) = \mathcal{N}(y_n|f(\bx_n), \sigma^2)$, which has an above form with the natural parameter $\lambda(y_n) = \{\frac{1}{\sigma^2}y_nk(\bx_n,\cdot), -\frac{1}{2\sigma^2}k(\bx_n, \cdot)k(\bx_n, \cdot)^\top\}$.
    By plugging in $p(y_n|f) \propto \exp\{\lambda(y_n)^\top t(f)\}$, we can verify that \cref{eq:natgrad-vi} is equivalent to
    \begin{equation*}
        q_{t+1}(f) \propto q_t(f)^{1 - \beta_t}p(f)^{\beta_t}p(y_n|f)^{N\beta_t},
    \end{equation*}
    which turns out to be the same adaptive Bayesian filter we get in \cref{eq:frac-bayes}. As for the non-conjugate case, we can view the natural gradient update as the projection of the functional mirror descent update onto exponential families, by approximating the likelihood term with the exponential family $\exp\{\langle\nabla_{u}\mathbb{E}_q[\log p(y_n|f)], t(f)\rangle\}$.
    
    \section{Experiment Details and Additional Results}
    \label{app:exp-details}
    
    \subsection{Synthetic Data} 
    
    \begin{figure*}[t]
        \centering
        \begin{subfigure}[t]{0.32\textwidth}
            \centering
            \includegraphics[width=\linewidth]{figures/toy/svgp-2}
        \end{subfigure}
        \begin{subfigure}[t]{0.32\textwidth}
            \centering
            \includegraphics[width=\linewidth]{figures/toy/svgp-5}
        \end{subfigure}
        \begin{subfigure}[t]{0.32\textwidth}
            \centering
            \includegraphics[width=\linewidth]{figures/toy/svgp-20}
        \end{subfigure} \\
        \begin{subfigure}[t]{0.32\textwidth}
            \centering
            \includegraphics[width=\linewidth]{figures/toy/gpnet-2}
        \end{subfigure}
        \begin{subfigure}[t]{0.32\textwidth}
            \centering
            \includegraphics[width=\linewidth]{figures/toy/gpnet-5}
        \end{subfigure}
        \begin{subfigure}[t]{0.32\textwidth}
            \centering
            \includegraphics[width=\linewidth]{figures/toy/gpnet-20}
        \end{subfigure} \\
        \begin{subfigure}[t]{0.32\textwidth}
            \centering
            \includegraphics[width=\linewidth]{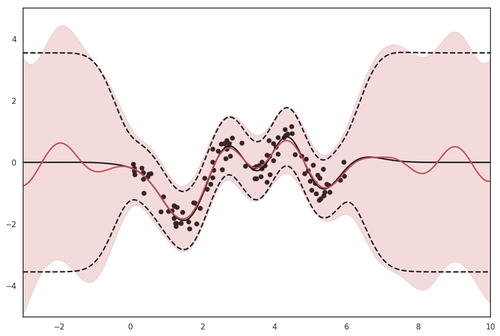}
        \end{subfigure}
        \begin{subfigure}[t]{0.32\textwidth}
            \centering
            \includegraphics[width=\linewidth]{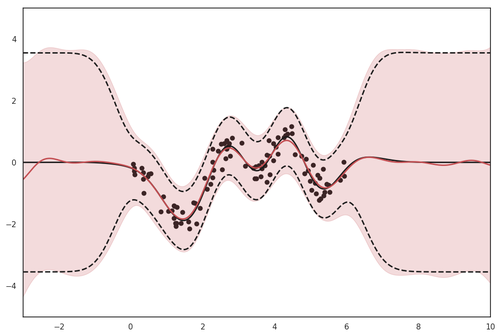}
        \end{subfigure}
        \begin{subfigure}[t]{0.32\textwidth}
            \centering
            \includegraphics[width=\linewidth]{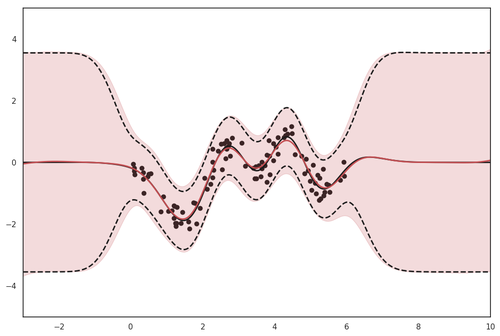}
        \end{subfigure}
        \caption{Posterior process on the Snelson dataset, where shaded areas correspond to intervals of $\pm3$ standard deviations, and dashed lines denote the ground truths. (a) From top to bottom: SVGP with $M\in\{2, 5, 20\}$ (left to right) inducing points; GPNet with $M\in\{2, 5, 20\}$ measurement points; FBNN with $M\in\{2, 5, 20\}$ measurement points.}
        \label{fig:snelson-ext}
    \end{figure*}
    
    The full figures including FBNN with $M=2, 5, 20$ are shown in Fig.~\ref{fig:snelson-ext}. %
    We can see that FBNN's problem of overestimating uncertainty is consistent with different values of $M$.
    We set $\beta_0 = 1, \xi=0.1$ in this experiment.
    
    \subsection{Regression}

    For all regression experiments, we set the measurement points to be sampled from the empirical distribution of training data convolved with the prior RBF kernel if the prior hyperparameters are initialized by optimizing GP marginal likelihood on a random subset, otherwise we set $c(\bx)$ simply to be the training data distribution. We fix $c(\bx)$ and do not adapt it together with the prior kernel parameters when the prior hyperparameters are updated during training.
    
    \paragraph{Benchmarks}The GP we use has a RBF kernel with dimension-wise lengthscales, %
    also known as \emph{Automatic Relevance Determination} (ARD)~\citep{mackay1996bayesian}. The RFE inference network we use has 1000 features (hidden units). We initialize the lengthscales in the network using the lengthscales of the prior kernel, and initialize the frequencies (first-layer weights) with random samples from a standard Gaussian. We then train these frequencies and lengthscales as inference network parameters.
    For FBNN, we keep all settings (including the inference network) the same as GPNet except the training objective used.
    We use 20 random splits for each dataset, where we keep 90\% of the dataset as the training set and use the remaining 10\% for test. The inputs and outputs of all data points are scaled to have nearly zero mean and unit variance using the mean and standard deviation calculated from training data. We use minibatch size 500, learning rate $\eta=0.003$ for all the experiments. With each random seed we ran 10K iterations. 
    For Boston, Concrete, Kin8nm and Protein we initialize the prior hyperparameters by maximizing GP marginal likelihood for 1K iterations on a randomly chosen subset of 1000 points.
    For Power and Wine we optimize the prior hyperparameters during training using the minibatch lower bound and the same learning rate as $\eta$, as described in \cref{sec:algo}. We set $\beta_0=1$ and $\xi=1$ except for Power and Protein we use $\beta_0=0.01, \xi=0.1$ and $\beta_0=0.1, \xi=0.1$, respectively.
    In \Cref{tab:rmse} and \Cref{tab:test-ll} we list the mean and standard errors of all experiments.
    
    \paragraph{Airline Delay} We use the same type of GP prior and inference networks as in benchmark datasets above. We set $\beta_0=0.1, \xi=0.1$. For all methods we train for 10K iterations with minibatch size 500 and learning rate $\eta=0.003$. For GPNet we optimize the prior hyperparameters during training with the same learning rate as $\eta$, for which the objective is described in \cref{sec:algo}, while we found that doing the same with FBNN seriously hurt its performance (leading to RMSE 27.186 for $M=100$), therefore we did not update hyperparameters during training for FBNN. We initialize the prior hyperparameters by maximizing GP marginal likelihood for 1K iterations on a randomly chosen subset of 1000 points. We did this for both GPNet and FBNN, though we found that for GPNet this does not improve the performance.
    
    \begin{table*}[t]
        \caption{Regression: RMSE.}
        \label{tab:rmse}
        \vskip 0.15in
        \begin{center}
            \begin{small}
                \begin{sc}
                    \begin{tabular}{lccrrrrr}
                        \toprule
                        Data set & N & D & \makecell{SVGP, 100} & \makecell{GPNet, 100} & \makecell{SVGP, 500} & \makecell{GPNet, 500} &
                        \makecell{FBNN, 500}\\
                        \midrule
                        Boston & 506 & 12 &  2.897$\pm$0.132 & 2.786$\pm$0.142 & 3.023$\pm$0.187 & 2.754$\pm$0.143 & 
                        2.775$\pm$0.141\\
                        Concrete & 1030 & 8 & 5.768$\pm$0.094 & 5.301$\pm$0.127 & 5.075$\pm$0.119 &  5.050$\pm$0.132 &
                        5.089$\pm$0.117 \\
                        Energy & 768 & 8 & 0.469$\pm$0.014 & 0.493$\pm$0.022 & 0.439$\pm$0.015 &  0.461$\pm$0.014 &
                        0.459$\pm$0.013 \\
                        Kin8nm & 8192 & 8 & 0.086$\pm$0.001 & 0.080$\pm$0.001 & 0.074$\pm$0.000& 
                        0.067$\pm$0.000 &
                        0.072$\pm$0.000 \\
                        Power & 9568 & 4 &  3.941$\pm$0.033 & 3.942$\pm$0.032 & 3.791$\pm$0.034 & 3.898$\pm$0.032 &
                        4.135$\pm$0.029 \\
                        Protein & 45730 & 9 & 4.536$\pm$0.010 & 4.540$\pm$0.014 & 4.154$\pm$0.010 & 4.329$\pm$0.013 &
                        4.087$\pm$0.051 \\
                        Wine & 1599 & 11 & 0.625$\pm$0.009 & 0.614$\pm$0.010 & 0.626$\pm$0.009 & 0.627$\pm$0.009 &
                        0.633$\pm$0.008 \\
                        \bottomrule
                    \end{tabular}
                \end{sc}
            \end{small}
        \end{center}
        \vskip -0.1in
    \end{table*}
    
    \begin{table*}[t]
        \caption{Regression: Test log likelihood.}
        \label{tab:test-ll}
        \vskip 0.15in
        \begin{center}
            \begin{small}
                \begin{sc}
                    \begin{tabular}{lccrrrrr}
                        \toprule
                        Data set & N & D & \makecell{SVGP, 100} & \makecell{GPNet, 100}  & \makecell{SVGP, 500} & \makecell{GPNet, 500} & \makecell{FBNN, 500}\\
                        \midrule
                        Boston & 506 & 12 &  -2.465$\pm$0.054 & -2.421$\pm$0.049 & -2.458$\pm$0.072 & -2.429$\pm$0.055 &
                        -2.437$\pm$0.025  \\
                        Concrete & 1030 & 8  & -3.166$\pm$0.015 & -3.115$\pm$0.024 & -3.027$\pm$0.023 & -3.066$\pm$0.022 &
                        -3.046$\pm$0.029 \\
                        Energy & 768 & 8  & -0.675$\pm$0.024 & -1.060$\pm$0.008 & -0.600$\pm$0.033 & -0.847$\pm$0.013 &
                        -0.755$\pm$0.018 \\
                        Kin8nm & 8192 & 8  & 1.006$\pm$0.004 & 1.095$\pm$0.011 & 1.183$\pm$0.004 & 1.283$\pm$0.005 &
                        1.189$\pm$0.005\\
                        Power & 9568 & 4 &  -2.793$\pm$0.008 & -2.794$\pm$0.007 & -2.755$\pm$0.008 & -2.783$\pm$0.008 &
                        -2.847$\pm$0.006 \\
                        Protein & 45730 & 9 & -2.932$\pm$0.002 & -3.057$\pm$0.032 & -2.841$\pm$0.002 & -2.986$\pm$0.029 &
                        -2.821$\pm$0.014\\
                        Wine & 1599 & 11 &  -0.949$\pm$0.014 & -0.917$\pm$0.014 & -0.949$\pm$0.015 & -0.948$\pm$0.014 &
                        -0.961$\pm$0.013\\
                        \bottomrule
                    \end{tabular}
                \end{sc}
            \end{small}
        \end{center}
        \vskip -0.1in
    \end{table*}
    
    \subsection{Classification}
    \label{app:exp-cls}
    
    \paragraph{CNN-GP Prior} The prior is defined as follows. Let $\bZ^{(\ell)}(\bx)$ denote the pre-activation output of the $\ell$-th layer of the ConvNet. The shape of $\bZ^{(\ell)}(\bx)$ is $C^{(\ell)}\times(H^{(\ell)}D^{(\ell)})$. Each row of it represents the flattened feature map in a channel. A hidden layer in the network makes the transformation:
    \begin{equation*}
    Z_{j, g}^{(\ell + 1)}(\bx) = b_j^{(\ell)} + \sum_{i=1}^{C^{(\ell)}}\sum_{h=1}^{H^{(\ell)}D^{(\ell)}}W_{j,i,g,h}^{(\ell)}a(Z_{i,h}^{(\ell)}(\bx)),
    \end{equation*}
    where $\bW_{ji}$ is the pseudo weight matrix that corresponds to the convolutional filter $\bU_{ji}$. The elements of each row in $\bW_{ji}$ are zero except where $\bU_{ji}$ applies. $b_j$ denotes the bias in the $j$-th channel. $a$ is the ReLU activation function. Let $x, y$ denote the positions within a filter, independent Gaussian priors are placed over $u_{j,i,x,y}^{(\ell)}$ and $b_j^{(\ell)}$ to form a Bayesian ConvNet: $$
    u_{j,i,x,y}^{(\ell)}\sim \mathcal{N}(0, \sigma_w^2/C^{(\ell)}),\quad b_j\sim \mathcal{N}(0,\sigma_b^2).$$ 
    By carefully taking the limit of hidden-layer widths, one can prove that each row in $\bZ^{(\ell)}(\bx)$ form a multivariate Gaussian, and different rows (channels) are independent and identically distributed (i.i.d.), thus showing that the Bayesian ConvNet defines a GP~\citep{garriga-alonso2018deep}.
    It is easy to show the prior mean function is zero: $
    \mathbb{E}[Z_{j,g}^{(\ell + 1)}(\bx)] = 0. $
    To determine the prior covariance kernel of the output, we can follow a recursive procedure 
    (For simplicity, we use $v_g^{(\ell)}(\bx, \bx')$ to denote the covariance between $Z_{j,g}^{(\ell)}(\bx)$ and $Z_{j,g}^{(\ell)}(\bx')$):
    \begin{align*}
    v_g^{(\ell + 1)}(\bx,\bx') &= \sigma_b^2 + \sigma_w^2\sum_{h\in g\text{-th patch}}s_h^{(\ell)}(\bx, \bx'), \\
    s_h^{(\ell)}(\bx, \bx') &= 1/{2\pi}\sqrt{v_g^{(\ell)}(\bx, \bx)v_g^{(\ell)}(\bx', \bx')}J_1(\theta_g^{(\ell)}),
    \end{align*}
    where $J_1(\theta_g^{(\ell)}) = \sin\theta_g^{(\ell)} + (\pi - \theta_g^{(\ell)})\cos\theta_g^{(\ell)}$ and $\theta_g^{(\ell)}= \arccos\left(v_g^{(\ell)}(\bx, \bx')/\sqrt{v_g^{(\ell)}(\bx, \bx)v_g^{(\ell)}(\bx',\bx')}\right)$. The prior convnet we used is a deep convolutional neural network with 6 residual blocks, each two of them operates on a different size of feature maps, with the first two on feature maps with the same size as the original image. There are strided convolution (stride=2) between the three groups.
    
    \paragraph{Inference Networks} The inference network we used has the same structure as the prior ConvNet, except the number of convolutional filters are [64, 64, 128, 128, 256, 256]. On top of it we have a fully-connected layer of size 512 and neural tangent kernels defined by a MLP with 100 hidden units for the output of each class. %
    
    We use batch size 64 and $M=64$ measurement points in this experiment. We set $c(\bx)$ to be the empirical distribution of the training data. In implementation this simply means that we use two different shuffles of the training dataset, and pick a minibatch from each of them. Then we use one of the two minibatches as training points, and the other as measurement points. The learning rate is $\eta=0.0003$. We set $\beta_0=0.01, \xi=0.1$, and ran for 10K iterations. We did not update prior hyperparameters in this experiment.

\end{document}